\documentclass[12pt,a4paper]{article}

\usepackage[numbers,sort&compress]{natbib}  
\usepackage[ruled,vlined]{algorithm2e}

\usepackage[a4paper,top=2cm,bottom=2cm,left=2.5cm,right=2.5cm,marginparwidth=1.75cm]{geometry}
\usepackage[parfill]{parskip}
\usepackage{multirow}

\usepackage{amsmath}
\usepackage{graphicx}
\usepackage[colorlinks=true, allcolors=blue]{hyperref}
\usepackage{cleveref}
\usepackage[title]{appendix}
\usepackage{mathrsfs}
\usepackage{amsfonts}
\usepackage{booktabs} 
\usepackage{caption}  
\usepackage{threeparttable} 
\usepackage{algorithmicx}
\usepackage{algpseudocode}
\usepackage{listings}
\usepackage{enumitem}
\usepackage{chngcntr}
\usepackage{booktabs}
\usepackage{lipsum}
\usepackage{subcaption}
\usepackage{authblk}
\usepackage[T1]{fontenc}    
\usepackage{csquotes}       
\usepackage{diagbox}
\usepackage{comment}

\usepackage{setspace}
\usepackage{titlesec}
\titleformat{\section} 
  {\normalfont\Large\bfseries}{\thesection.}{1em}{}
  
\usepackage{lineno} 

\rightlinenumbers 

\usepackage{float}   
\usepackage{caption} 

\title{Efficiently Training Large Neural Networks With Low-Dimensional Error Feedback}

\author{Maher Hanut}
\author{Jonathan Kadmon}
\affil{\small Edmond and Lily Safra Center for Brain Sciences, \\
The Hebrew University,\\
Jerusalem.\\
\texttt{\{maher.hanut, jonathan.kadmon\}@mail.huji.ac.il}}

\date{}  
\begin{document}
\maketitle

\begin{abstract}
Training large neural networks relies on propagating detailed error signals backward through every layer, a procedure that is computationally expensive and difficult to reconcile with biological teaching pathways. In the brain, instructive signals are typically sparse, indirect, and compressed, raising the question of whether a low-dimensional teaching signal can suffice for effective credit assignment without transmitting a full layerwise gradient. Building on Feedback Alignment, we introduce a low-dimensional feedback framework that decouples the backward and forward passes and explicitly controls the dimensionality of the error signal delivered to each layer while preserving high-dimensional feedforward representations. We develop a theory for the nonlinear learning dynamics induced by gradient descent in linear networks. Fixed, task-agnostic low-dimensional feedback can misassign credit and lead to spurious solutions, whereas allowing the feedback subspace to adapt toward task-relevant error directions recovers backpropagation-like learning dynamics. Guided by these predictions, we train convolutional networks and vision transformers to backpropagation-level performance with modest feedback dimensionality that scales with task dimensionality rather than network width, while reducing total training compute by over 20\% without loss of accuracy. We further derive synapse-local rules that learn the relevant error subspace from pre- and postsynaptic signals, providing a biologically grounded mechanism for compressed teaching. Finally, feedback dimensionality reshapes receptive fields, revealing the teaching pathway as an inductive bias in representation learning.
\end{abstract}



\newpage
\paragraph*{Significance statement}
Training modern neural networks often depends on sending detailed error information backward through every layer, a process that is costly in time, energy, and memory. In contrast, the brain appears to rely on simpler teaching signals. We show that accurate learning can be driven by a compact feedback signal whose size is set by task demands rather than by the number of neurons or synapses. Building on this principle, we develop a training framework that matches the accuracy of the standard training method while being more efficient. We also describe local learning rules for constrained teaching pathways and show that these constraints shape features learned in early layers, offering a new route for modeling learning in brain circuits.

\section{Introduction}\label{sec:introduction}
Training large neural networks depends on assigning credit across many layers and millions of parameters. The standard approach, gradient descent with error backpropagation, computes reverse-mode derivatives by transporting high-dimensional error information through every layer. This backward pass is not only a conceptual constraint but also a practical bottleneck because it requires manipulating and storing large intermediate signals whose size scales with hidden-layer width, leading to substantial memory, bandwidth, and compute costs during training \cite{baydin2018autodiff,chen2016sublinear,rhu2016vdnn}. While large forward representations are needed for expressivity, it is not obvious that the feedback pathway must also scale with model size.

This raises a basic efficiency question. Many tasks have low-dimensional outputs, so the loss gradient at the output layer is naturally low-dimensional, with an effective dimensionality tied to task complexity rather than to hidden-layer width. The open issue is whether deep networks can be trained by transmitting only a task-relevant, low-dimensional teaching signal to each layer, thereby reducing the overhead of the backward pass without sacrificing the benefits of high-dimensional feedforward representations.

This efficiency-motivated perspective is also consistent with biological circuitry. Biological circuits do not exhibit the synapse-wise, bidirectional connectivity and precise weight symmetry required by gradient backpropagation. Instead, experimentally observed instructive channels are typically indirect, not originating from the direct downstream neural population, and are often sparse and long-range, which implies a restricted teaching signal~\cite{marder2012neuromodulation}. Notable examples include midbrain dopaminergic neurons~\cite{Schultz1997Neural,Montague1996Framework,bayer2005midbrain}, the olivocerebellar system \cite{ito2016error,ohmae2015climbing}, cholinergic neuromodulation~\cite{kilgard1998cortical,froemke2007synaptic,letzkus2011disinhibitory}, and the noradrenergic system in the Locus Coeruleus~\cite{aston2005integrative,bouret2005network,sara2012orienting,uematsu2015projection,chandler2019redefining}. Combined with empirical evidence that task-relevant population activity is often confined to low-dimensional manifolds~\cite{churchland2012neural,cunningham2014dimensionality,sadtler2014neural,gao2015simplicity}, these observations suggest that the brain’s teaching pathway is more plausibly indirect and compressed (low-dimensional) than a neuron-by-neuron gradient trace.

A parallel question arises in artificial learning. Existing biologically motivated alternatives to backpropagation, such as Feedback Alignment (FA) \cite{lillicrap2016random}, decouple feedforward and feedback weights but still propagate full-dimensional error signals and often fall short of backpropagation in deep or complex architectures \cite{bartunov2018assessing}. As a result, the role of error dimensionality as an explicit constraint on deep credit assignment has been difficult to isolate.

Here we develop a theoretical and algorithmic framework for training large networks with constrained, low-dimensional error feedback. We treat the error pathway as a controllable resource and ask which error subspace must be transmitted for effective learning, and how such a subspace can be learned with local plasticity rules. Our approach imposes a rank-$r$ constraint on feedback and learns a task-relevant low-dimensional teaching channel, thereby decoupling the backward pathway from the feedforward computation while preserving high-dimensional forward representations. We refer to this framework as low-dimensional feedback alignment (LDFA).

Our work makes four contributions:
\begin{itemize}
\item We show that accurate credit assignment does not require full-dimensional backward signals. In a solvable linear theory, fixed low-rank feedback makes the stationarity conditions underdetermined and introduces spurious fixed points, and we characterize how the rank required for backpropagation-like convergence depends on task dimensionality.
\item We introduce a factorized low-rank feedback pathway together with learning rules that make the dimensionality of the layerwise teaching signal an explicit, tunable resource.
\item We demonstrate in deep nonlinear networks, including convolutional architectures and vision transformers, that near-backpropagation accuracy is achievable with ranks on the order of task complexity and largely independent of network width, and we quantify the resulting reduction in backward-pass compute.
\item We show that error dimensionality systematically shapes learned representations, including receptive-field structure in a simplified ventral-stream model.
\end{itemize}

Together, these results challenge the assumption that scalable credit assignment must involve transporting a full high-dimensional gradient through depth. Instead, they support a view in which learning can be driven by a learned low-dimensional teaching channel that preserves task-relevant error directions, yielding a computationally efficient training regime and a concrete hypothesis for how anatomically restricted feedback pathways could still support deep learning.

\section{Background and related work}
\label{sec:background}
We consider a multilayered perceptron with $L$ layers, each layer $l$ computes its output as $\boldsymbol{h}_l = f(W_l \boldsymbol{h}_{l-1})$, where $W_l$ is the weight matrix and $f$ is an element-wise activation function. The input to the network is $\boldsymbol{h}_0 = \boldsymbol{x}$, and the final network output is $\hat{\boldsymbol{y}} = f_L(W_L \boldsymbol{h}_{L-1})$, which approximates the target $\boldsymbol{y}$. The \textit{task dimensionality}, denoted $d$, is at most the number of components in $\boldsymbol{y}$ and $\hat{\boldsymbol{y}}$.
Training the network involves minimizing a loss function $\mathcal{L}(\boldsymbol{y}, \hat{\boldsymbol{y}})$ by adjusting the weights $\{W_l\}$. The error signal at the output layer, $\delta_L = \frac{\partial \mathcal{L}}{\partial \hat{\boldsymbol{y}}}$, is a $d$-dimensional vector, typically much smaller than the number of neurons in the hidden layers.

\textbf{Backpropagation (BP)}~\cite{rumelhart1986learning} is the standard approach for training neural networks. It propagates the error backward through the network using $\delta_l = W_{l+1}^\top \delta_{l+1} \odot f'(W_l \boldsymbol{h}_{l-1})$, and updates the weights using $\Delta W_l = -\eta \delta_l \boldsymbol{h}_{l-1}^\top$, where $\eta$ is the learning rate. However, this method requires an exact transposition of the forward weights, $W_{l+1}^\top$, which is biologically implausible~\cite{grossberg1987competitive, crick1989recent}. Moreover, BP tightly couples the error propagation with the forward pass, limiting the ability to explore how different properties of the error signal affect learning dynamics.

\textbf{Feedback Alignment (FA)}~\cite{lillicrap2016random} was proposed to address the biological limitations of BP by replacing $W_{l+1}^\top$ with a fixed random matrix $B_l$. The error is computed as:
\begin{equation}\label{eq:FAerror}
    \delta_l = B_l \delta_{l+1} \odot f'(W_l \boldsymbol{h}_{l-1}),
\end{equation}
decoupling the forward and backward weights and providing a more biologically plausible mechanism. However, FA struggles to scale effectively in deep architectures, such as convolutional neural networks (CNNs), where it often fails to achieve competitive performance~\cite{bartunov2018assessing}.

An extension of FA involves adapting $B_l$ by updating it alongside the forward weights $W_l$ to improve their alignment~\cite{kolen1994backpropagation, akrout2019deep}:
\begin{equation}\label{eq:FAupdate}
    \Delta B_l = -\eta \boldsymbol{h}_{l-1} \delta_l^\top - \lambda B_l, \quad
    \Delta W_l = -\eta \delta_l \boldsymbol{h}_{l-1}^\top - \lambda W_l,
\end{equation}
where $\lambda$ is a regularization parameter. Although this adaptive approach improves performance by better aligning forward and backward weights, it still requires high-dimensional error signals and and is less efficient than backpropagation. Furthermore, in  \cref{sec:linear}, we show that this approach fails when the matrix $B$ is low-rank and the dimensionality of the error is constrained.

Other studies have explored the use of \textit{fixed} sparse feedback matrices to reduce the dimensionality of error propagation~\cite{crafton2019direct}. However, these approaches result in significantly lower performance and do not provide a systematic framework for studying how error constraints affect learning and representation formation.

Beyond FA-based methods, several studies have shown that weight updates using backpropagation can result in a low-dimensional weight update~\cite{liao2016important,gunasekar2018implicit, caro2024translational} and favor low-rank solutions~\cite{patel2024learning}.  These findings support our hypothesis that low-dimensional feedback is sufficient to train deep networks. However, no previous work has considered training with a constrained error pathway, and the effects of error dimensionality and training have not been systematically studied.

\begin{figure}
    \centering
    \includegraphics[width=1.0\linewidth]{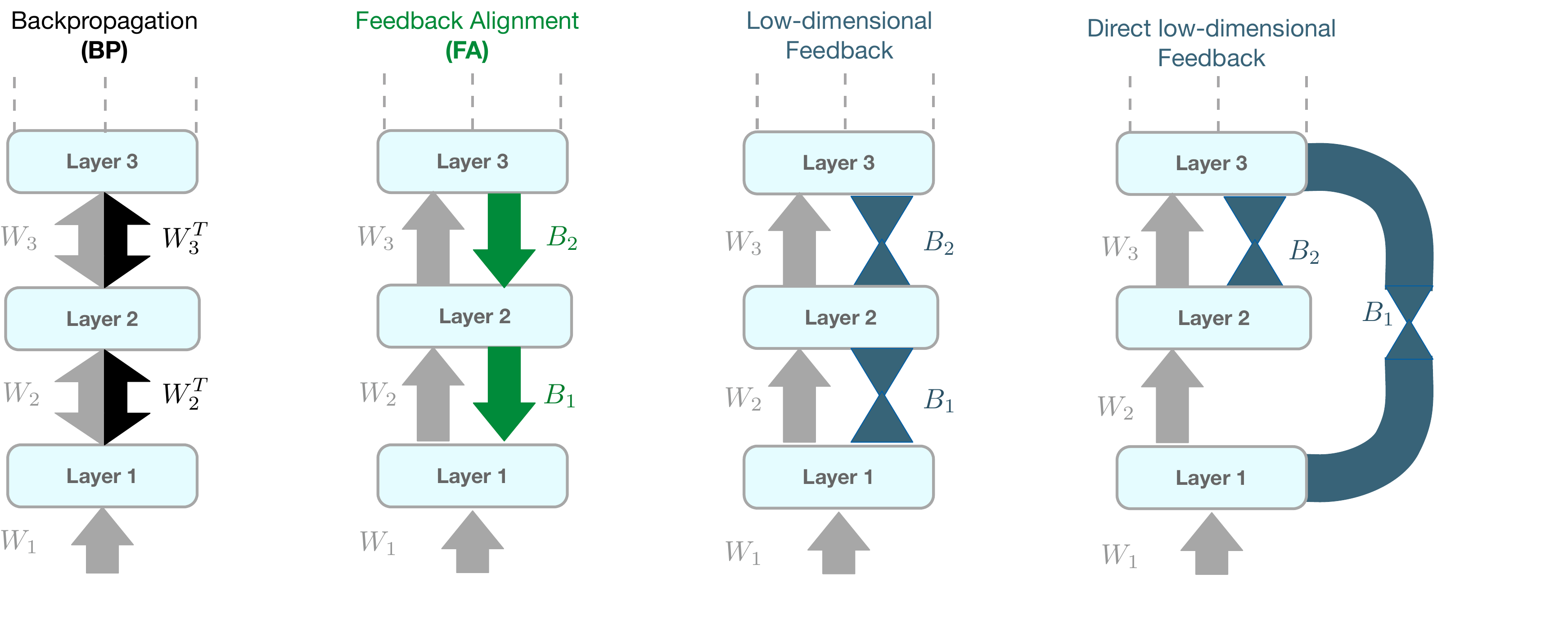}
  \caption{\textit{Illustration of different approaches for propagating error to hidden layers.} From left to right: \textbf{Backpropagation (BP)} propagates error using the exact transposes of the forward weights. \textbf{Feedback alignment (FA)} replaces $W_l^\top$ with fixed random feedback matrices and relies on the forward weights aligning with this backward pathway during training. \textbf{Low-dimensional feedback alignment (LDFA)} constrains each feedback map to be low rank (e.g., $B_l=Q_lP_l$ with rank $r$), so each layer receives an $r$-dimensional teaching signal whose subspace is learned. \textbf{Direct LDFA (dLDFA)} extends LDFA by routing low-dimensional error projections to non-adjacent layers, allowing early layers to receive teaching signals directly from the output (or other higher) layers.}

    \label{fig:schematics}
\end{figure}

In this work, we ask how limiting the dimensionality of the \emph{propagated error} constrains credit assignment in large neural networks. We treat the error pathway as a controllable resource and study how its  dimensionality shapes learning dynamics, computational cost, and the representations that emerge during training. To do so, we introduce a learning scheme in which each layer receives error through a low-rank feedback map, enabling explicit control over the dimensionality of the layerwise teaching signal (\cref{fig:schematics}).

This focus differs from prior observations that learning dynamics or weight updates often exhibit low-dimensional structure: those results describe emergent compression but do not impose a bottleneck on the communicated error itself. The remainder of the paper develops a solvable linear theory of low-rank error feedback (\cref{sec:linear}), tests the predicted scaling in deep vision models (\cref{sec:deep_nonlinear}), derives synapse-local rules for learning the feedback pathway (\cref{sec:local_rule}), and then analyzes efficiency and representational consequences before closing with broader implications in the Discussion.

\section{Low-dimensional error feedback in linear networks}
\label{sec:linear}
We begin our analysis by studying learning dynamics in multilayered linear networks. Although linear models may seem overly simplistic, they can exhibit rich learning dynamics due to the nonlinearity introduced by the loss function~\cite{saxe2013exact}. Additionally, imposing dimensional constraints on linear networks yields insightful results that extend beyond the linear case.

\paragraph{A linear problem}
We consider a simple linear transformation problem with a low-dimensional structure, $\boldsymbol{y} = A\boldsymbol{x}$. Here, $\boldsymbol{x} \in \mathbb{R}^n$ represents the $n$-dimensional input, and $\boldsymbol{y} \in \mathbb{R}^m$ represents the target. The matrix $A$ is a rank-$d$ matrix defined as $A = \sum_{j=1}^d \boldsymbol{u}_j \boldsymbol{v}_j^\top$, where $\boldsymbol{u}_j \in \mathbb{R}^n$ and $\boldsymbol{v}_j \in \mathbb{R}^m$ are random Gaussian vectors, and we assume $d \ll n$. Our data set consists of $p$ training samples $\{\boldsymbol{x}^\mu, \boldsymbol{y}^\mu\}_{\mu=1}^p$, with each input vector $\boldsymbol{x}^\mu$ being i.i.d. according to the standard normal distribution $\boldsymbol{x}^\mu \sim \mathcal{N}(\boldsymbol{0}, \boldsymbol{1})$. The labels are given by $\boldsymbol{y}^\mu = A\boldsymbol{x}^\mu + \boldsymbol{\xi}^\mu$, where $\boldsymbol{\xi}^\mu$ is additive Gaussian noise with zero mean and unit variance. 

The goal is to learn the low-dimensional structure of $A$ from the $p$ samples using a linear neural network. For simplicity, we assume that $p$ is large enough to allow the network to fully recover the structure of $A$.

\paragraph{A linear network model.}
To study the effects of restricted error pathways, we consider a simple linear network with three layers: an input layer $\boldsymbol{x} \in \mathbb{R}^n$, a hidden layer $\boldsymbol{h} \in \mathbb{R}^k$, and an output layer $\boldsymbol{y} \in \mathbb{R}^m$. The input and hidden layers are connected by the weight matrix $W_1 \in \mathbb{R}^{k \times n}$, and the hidden and output layers are connected by the weight matrix $W_2 \in \mathbb{R}^{m \times k}$. The output of the network can be expressed as $\boldsymbol{y} = W_2 W_1 \boldsymbol{x}$ (Figure \ref{fig:linear-fig}).

The network is trained to minimize the quadratic empirical loss function:
\begin{equation}
    L = \frac{1}{p} \sum_\mu \|\boldsymbol{y}(\boldsymbol{x^\mu}) - W_2W_1 \boldsymbol{x}^\mu\|^2.
\end{equation}

We apply Feedback Alignment (FA) to update $W_1$, which does not have direct access to the loss gradient. Instead of backpropagating the error through $W_2^\top$, we use a fixed low-rank feedback matrix $B$.  This provides an alternative pathway for propagating the error signal to $W_1$.

For a given data point $\{\boldsymbol{x}^\mu, \boldsymbol{y}^\mu\}$, the weight updates, derived from the FA framework, are given by:
\begin{equation}\label{eq:DeltaW}
\begin{aligned}
    \Delta W_1^\mu& = \eta B (\boldsymbol{y}^\mu - W_2 W_1 \boldsymbol{x}^\mu) \boldsymbol{x}^{\mu \top }, \text{ and }\\
    \Delta W_2^\mu& = \eta (\boldsymbol{y}^\mu - W_2 W_1 \boldsymbol{x}^\mu) \boldsymbol{x}^{\mu \top } W_1^\top.
    \end{aligned}
\end{equation}
Here, $\eta$ represents the learning rate, and the update for $W_1$ is computed using the indirect error feedback provided by $B$, while $W_2$ receives the full error signal directly from the output.

\paragraph{Constraining error dimensionality with low-rank feedback}
To control the dimensionality of the error feedback, we impose a low-rank constraint on the feedback matrix $B$. Rather than allowing full-dimensional feedback, we decompose $B$ as $B = QP$, where $Q \in \mathbb{R}^{k \times r}$ and $P \in \mathbb{R}^{r \times m}$. When $r < \min(k, m)$, $B$ is low rank, which means that it can project the error signal onto at most $r$ independent directions.

This low-rank structure introduces an "$r$-bottleneck," which limits the flow of error information. In linear settings, the problem is solvable using a single weight matrix, rendering the training of \( W_1 \) unnecessary. However, backpropagation dynamics still adjust these weights~~\cite{saxe2013exact}. By controlling the value of $r$, we can systematically study how reducing the dimensionality of the error signal impacts learning. Initially, we follow the original Feedback Alignment framework, keeping $Q$ and $P$ as random matrices. However, as we will demonstrate, allowing $Q$ and $P$ to learn is crucial to high performance.

\subsection{Learning dynamics}
Our analysis extends the framework established by Saxe et. al.~\cite{saxe2013exact} to incorporate indirect feedback with constrained dimensionality.  We begin by characterizing the task across the $p$ data points through the input-output covariance matrix, $\Sigma_{io} = \frac{1}{p} \sum_{\mu=1}^p \boldsymbol{y}^\mu (\boldsymbol{x}^\mu)^\top$, which captures the correlation between input vectors $\boldsymbol{x}$ and output vectors $\boldsymbol{y}$. Performing Singular Value Decomposition (SVD), we obtain $\Sigma_{io} = USV^\top$, where $U \in \mathbb{R}^{m \times m}$ and $V \in \mathbb{R}^{n \times n}$ contain the left and right singular vectors, respectively, and $S \in \mathbb{R}^{m \times n}$ is a rectangular diagonal matrix of singular values. For sufficiently large $p$, the first $d$ singular values in $S$ correspond to the prominent directions in the data (i.e., the singular values of $A$), while the remaining singular values are $O(1/\sqrt{p})$ and dominated by noise.

To track how training aligns the network weights with these prominent directions, we rotate the weight matrices $W_1$, $W_2$, and $B$ using the singular vectors of $\Sigma_{io}$. This transformation simplifies the analysis by aligning the network's weight dynamics with the key data directions:
\[
W_1 = \bar{W}_1 V^\top, \quad W_2 = U \bar{W}_2, \quad B = \bar{B} U^\top,
\]
where $\bar{W}_1$, $\bar{W}_2$, and $\bar{B}$ represent the transformed weight matrices. This rotation aligns the weight dynamics with the dominant singular vectors, allowing us to focus on how the network captures the important features of the data.

Since the inputs are uncorrelated, we can apply these transformations to the iterative weight-update equations derived from the FA learning rule. Assuming a small learning rate $\eta \ll 1$ with full-batch updates, we express the weight updates in continuous time:
\begin{equation}\label{eq:Wdynamics}
    \tau \frac{d\bar{W}_1}{dt} = \bar{B} (S - \bar{W}_2 \bar{W}_1), \quad \tau \frac{d\bar{W}_2}{dt} = (S - \bar{W}_2 \bar{W}_1) \bar{W}_1^\top,
\end{equation}
where $\eta = dt / \tau$. This continuous form captures the dynamics of the learning process, allowing us to study it from a dynamical systems perspective. By analyzing these equations, we can identify fixed points and evaluate their stability, providing insight into how the network converges and learns under constrained feedback. 
\begin{figure}
    \centering
    \includegraphics[width=0.95 \linewidth]{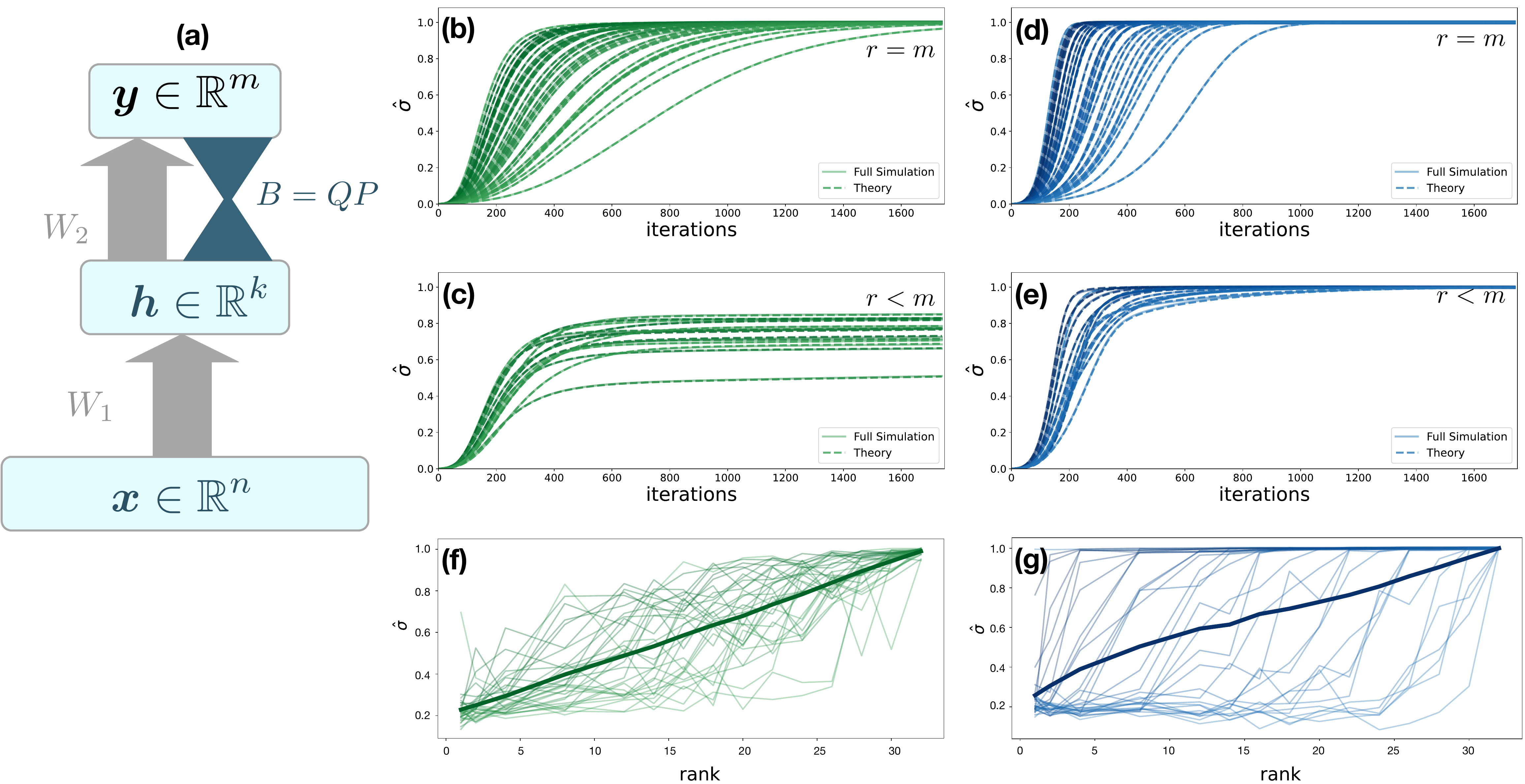}
 \caption{\textit{Learning dynamics and component alignment in linear networks.} 
\textbf{(a)} Schematic of the network architecture with input dimension $n = 128$, hidden layer size $k = 64$, and output dimension $m = 64$. The feedback matrix $B$ is factorized as $B = QP$ and constrained to rank $r$. 
\textbf{(b, c)} Comparison of theoretical predictions (dashed) and numerical simulations (solid) for low-rank Feedback Alignment (FA), updating $Q$ with $P$ fixed \cite{akrout2019deep}, with $r = m = 64$ and $r = 8 < m$, respectively. The $y$-axis reports the normalized mode overlap $\Lambda_i = u_i^\top W_2 W_1 v_i / S_{ii}$ for each singular component $i$. 
\textbf{(d, e)} Same as (b, c) but training both $Q$ and $P$ using Eqs.~\eqref{eq:BackwardUpdates}. 
\textbf{(f)} With $P$ fixed, overlaps increase on average (bold), but the leading singular components do not reliably reach $\Lambda_i \approx 1$ at low rank. 
\textbf{(g)} Training $P$ aligns the feedback subspace with the evolving error, yielding near-complete recovery of the top-$r$ components and improved convergence.}

    \label{fig:linear-fig}
\end{figure}
\paragraph{Stationary solutions for training.}
Training halts when the right-hand side of the weight dynamics in \eqref{eq:Wdynamics} vanishes, that is, when the system reaches a stationary point. At such a point, the update equation for $\bar{W}_1$ yields
\begin{equation} \label{eq:FixedpointCondition}
    \bar{B}\,(S - \bar{W}_2 \bar{W}_1) = 0
    \iff
    S_{jj}\,\boldsymbol{B}_{:,j} = \sum_{i=1}^m \boldsymbol{B}_{:,i}\,(\bar{W}_2 \bar{W}_1)_{ij}
    \quad \forall j,
\end{equation}
where $\boldsymbol{B}_{:,j}\in\mathbb{R}^k$ is the $j$-th column of $\bar{B}$ and $S_{jj}$ is the $j$-th singular value of $\Sigma_{io}$. Equation \eqref{eq:FixedpointCondition} constrains the mismatch between the target map $S$ and the learned input--output map $\bar{W}_2\bar{W}_1$ only through the subspace spanned by the columns of $\bar{B}$.

A convenient way to quantify how well the learned map captures each singular mode of the data is to project $\bar{W}_2\bar{W}_1$ into the singular-vector basis of $\Sigma_{io}=USV^\top$. We therefore define the (diagonal) mode overlap
\begin{equation}
    \Lambda_i
    = \frac{u_i^\top(\bar{W}_2\bar{W}_1)v_i}{S_{ii}}
    = \frac{(U^\top \bar{W}_2\bar{W}_1 V)_{ii}}{S_{ii}},
\end{equation}
where $u_i$ and $v_i$ are the $i$-th columns of $U$ and $V$. The overlap $\Lambda_i$ is a normalized measure of how strongly the learned input--output map expresses singular mode $i$ of $\Sigma_{io}$, with $\Lambda_i=1$ corresponding to perfect recovery of that mode.

When $\mathrm{rank}(\bar{B})=r=m$, the feedback pathway constrains all $m$ directions, and the stationary conditions select the backpropagation solution $\bar{W}_2\bar{W}_1=S$, as in the exact linear-network dynamics of Saxe et al.~\cite{saxe2013exact} (Fig.~\ref{fig:linear-fig}b). 

Crucially, when $r<m$, the constraint \eqref{eq:FixedpointCondition} is generically underdetermined because $\bar{B}$ can span at most $r$ independent directions. As a result, there can be infinitely many stationary solutions, and $\bar{W}_2\bar{W}_1$ need not match the full singular structure of $\Sigma_{io}$, so that some mode overlaps $\Lambda_i$ can deviate substantially from $1$ (Fig.~\ref{fig:linear-fig}c).


The implications of underdetermined fixed-point solutions are surprising. Naively, one would expect that if the bottleneck in the feedback is not too narrow, the projections of the error would maintain the necessary structure to guide the learning. Specifically, the Johnson–Lindenstrauss lemma~~\cite{johnson1984extensions} suggests that as long as $r>d\log m$, the pairwise correlation structure of the error signal is maintained. Nevertheless, our analysis shows that the solution weights are not guaranteed to converge to the correct solution. Thus, in the case of low-rank feedback, Feedback Alignment (FA) is likely to fail. Furthermore, FA fails when there is a bottleneck, i.e.,  $k<m,n$, even if the rank of $B$ is not restricted. The solution to this predicament, as we show next, is aligning the feedback weights with the data, ensuring that the network learns the correct representations.

\subsection{Training the feedback weights}
As shown above, when $\mathrm{rank}(B)=r<m$ the stationarity condition in Eq.~\eqref{eq:FixedpointCondition} is underdetermined, and fixed low-rank feedback (e.g., random $B$) can drive the dynamics to incorrect fixed points. We therefore keep the rank constraint, but allow the feedback pathway to adapt by parameterizing $B=QP$ with $Q\in\mathbb{R}^{k\times r}$ and $P\in\mathbb{R}^{r\times m}$.

The target feedback map is not known \textit{a priori}, but backpropagation provides an ideal reference: exact credit assignment is obtained for $B=W_2^\top$. We thus learn the best rank-$r$ approximation of $W_2^\top$ by minimizing the factorization loss
\begin{equation}
    \mathcal{L}_{B}=\frac{1}{2}\|QP-W_2^\top\|_F^2 .
    \label{eq:BackwardLoss}
\end{equation}
Notably, $\mathcal{L}_B$ involves only $W_2$ and the feedback parameters, so it is independent of the data and does not scale with batch size.

Gradient descent on $\mathcal{L}_B$ yields
\begin{equation}
\begin{aligned}[c]
\Delta P &= Q^\top\bigl(W_2^\top-QP\bigr),\\
\Delta Q &= \bigl(W_2^\top-QP\bigr)P^\top .
\end{aligned}
\label{eq:BackwardUpdates}
\end{equation}
Thus, the feedback updates do not require access to activities or error signals; their only coupling to the training problem is through the evolving forward weights $W_2$.

As in the analysis above, we rotate into the singular basis of $\Sigma_{io}$ by writing $P=\bar P U^\top$ and $W_2=U\bar W_2$, where $U$ contains the left singular vectors of $\Sigma_{io}$. Taking the continuous-time limit (with learning rate $\eta_B=dt/\tau_B$) gives the mode-wise dynamics
\begin{equation}
\begin{aligned}[c]
\tau_B\frac{d\bar P}{dt} &= Q^\top\bigl(\bar W_2^\top-Q\bar P\bigr),\\
\tau_B\frac{dQ}{dt}      &= \bigl(\bar W_2^\top-Q\bar P\bigr)\bar P^\top .
\end{aligned}
\label{eq:BackwardDynamics}
\end{equation}

Together with the forward dynamics, Eqs.~\eqref{eq:BackwardDynamics} define a closed system in which feedforward and feedback weights co-adapt. As shown in \cref{fig:linear-fig}c,d, learning both $P$ and $Q$ eliminates the spurious low-rank fixed points and restores convergence to the correct solution under constrained feedback.

\section{Low-dimensional error in deep vision models}
\label{sec:deep_nonlinear}
The linear theory above isolates the key obstruction induced by fixed low-rank feedback: when $\mathrm{rank}(B)=r<m$, the stationarity condition becomes underdetermined and learning can converge to spurious fixed points. Learning the feedback weights removes this degeneracy by making the backward mapping track the evolving feedforward transposes. Here we show that the same mechanism supports effective training in modern deep vision architectures, and we quantify the resulting scaling and computational consequences.

This extension is conceptually direct. The feedback factors are updated only through the mismatch $|Q_lP_l-W_l^\top|_F^2$, which depends on the feedforward weights but not on activities, batch statistics, or the sampled errors used to train $W_l$. Consequently, inserting local nonlinearities in the forward pass does not alter the feedback-learning rule. We first state the multilayer nonlinear construction explicitly, and then apply it to convolutional and transformer-based vision models.

\subsection{Multilayer nonlinear perceptrons}
We consider an $L$-layer multilayer perceptron with activities $\boldsymbol{h}0=\boldsymbol{x}$ and
\begin{equation}
\boldsymbol{a}_{l+1}=W_l\boldsymbol{h}l,\qquad \boldsymbol{h}_{l+1}=f(\boldsymbol{a}_{l+1}),\qquad l=0,\ldots,L-1,
\end{equation}
where $f$ is applied element-wise. Let $\boldsymbol{\delta}_l$ denote the backpropagated error at layer $l$, including the derivative of the nonlinearity. Under low-dimensional feedback, we replace the transpose $W_l^\top$ in the backward recursion with a learned rank-$r$ factorization $B_l=Q_lP_l$, giving
\begin{equation}
\boldsymbol{\delta}_l=(Q_lP_l\boldsymbol{\delta}_{l+1})\odot f'(\boldsymbol{a}_l),\qquad l=1,\ldots,L-1,
\end{equation}
with the usual initialization at the output layer determined by the loss.

Given these error signals, the feedforward weights follow the standard local outer-product update,
\begin{equation}\label{eq:Wlupdate}
\Delta W_l^\mu=\eta\boldsymbol{\delta}_{l+1}^\mu\boldsymbol{h}_l^{\mu\top}-\lambda W_l,
\end{equation}
where $\lambda$ is an $L_2$ regularization (weight decay) parameter.

Finally, the feedback factors are trained to track the evolving transposes by gradient descent on $\frac{1}{2}|Q_lP_l-W_l^\top|_F^2$, yielding
\begin{equation}
\begin{aligned}[c]
\Delta P_l=&Q_l^\top\big(W_l^\top-Q_lP_l\big),\quad \text{and}\\
\Delta Q_l=&\big(W_l^\top-Q_lP_l\big)P_l^\top.
\end{aligned}
\end{equation}
These updates depend only on $W_l$ and can therefore be computed without storing activations or using training data; in practice, they may also use a separate step size and can be applied on a slower update schedule than the forward weights.

\subsection{Deep convolutional networks}
Training convolutional networks with Feedback Alignment (FA) is known to be challenging~\cite{bartunov2018assessing, launay2019principled}. Recent work improves FA by learning feedback weights~\cite{bacho2024low}, but our approach leverages a low-rank parameterization $B=QP$ that adapts both the output subspace that carries the propagated error (through $Q$) and the input subspace from which that error is assembled (through $P$). This factorized learning rule makes it possible to test whether deep convolutional networks can be trained with an error signal whose dimension is far smaller than the feedforward width, here measured by the number of channels per layer.

We trained a VGG-like convolutional network~\cite{simonyan2014very} with four blocks and batch normalization on CIFAR-10 (see Appendix for details). In all layers, we replaced transpose feedback with learned low-rank backward matrices. Within each block, we fixed the feedback rank to be a fraction of the block width: $1/2$, $1/4$, or $1/8$ (\cref{fig:nonlinear}a). Reducing the error dimensionality had little effect on test performance, except in the most extreme setting. When the rank was $1/8$ of the block width, performance dropped noticeably, consistent with the linear prediction that accuracy degrades when $r<d$: in our smallest block (64 channels), rank-$1/8$ feedback corresponds to $r=8$, below the CIFAR-10 task dimensionality $d=10$. Overall, these results show that deep convolutional networks can be trained effectively using a strongly compressed error signal.

\subsection{Transformer-based architectures}
Transformers achieve state-of-the-art performance in image classification~\cite{dosovitskiy2020image,khan2022transformers}. Unlike convolutional and standard feedforward networks, they rely on self-attention, which mixes token representations through multiplicative interactions between key, query, and value projections~\cite{vaswani2017attention}. These interactions introduce strong cross-token dependencies, so it is not \textit{a priori }clear that strongly compressed error signals provide sufficient credit assignment for learning in transformer models.

To test this, we applied our low-rank feedback scheme to a vision transformer trained on CIFAR-10. All weight matrices in the architecture, including the key, query, and value projections, were trained using low-dimensional error feedback from the subsequent layer (see Appendix for details). For simplicity, we used the same feedback rank for all matrices, so the error dimensionality is controlled solely by the rank $r$ of each learned backward map. Performance as a function of feedback rank is shown in \cref{fig:nonlinear}b.

Despite the complexity of self-attention, transformers remained trainable with error dimensionality as low as the task dimensionality. In particular, rank $r=24$ feedback yielded performance on CIFAR-10 comparable to standard backpropagation (Figure~\ref{fig:nonlinear}b,c), consistent with our observations in linear and convolutional networks.

\subsection{Scaling and efficiency gains}
The results above show that low-dimensional feedback suffices to train deep nonlinear architectures, including convolutional networks and transformers, to near backpropagation-level accuracy. This has two implications: it decouples how performance scales with feedforward width from how the feedback pathway must scale, and it opens a regime where training can be more computationally efficient than standard backpropagation.

\begin{figure}
    \centering
    \includegraphics[width=1.0 \linewidth]{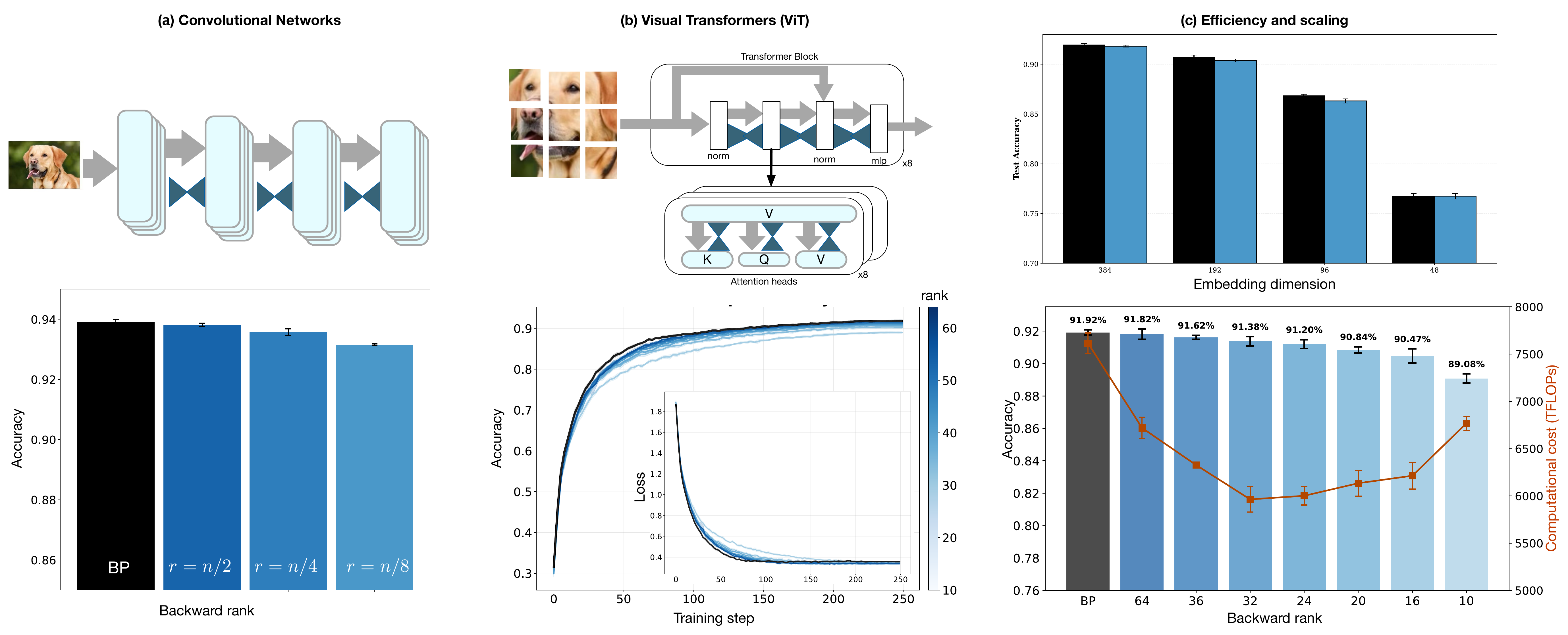}
   \caption{\textit{Low-dimensional feedback in nonlinear vision architectures.} \textbf{(a)} Convolutional networks. A VGG-like CNN trained on CIFAR-10 with learned low-rank feedback achieves near-backpropagation (BP) test accuracy even when the backward/teaching channel is strongly compressed. Bars sphow final accuracy for BP and for low-dimensional feedback with feedback rank set to a fraction of the layer width ($r=n/2,\,n/4,\,n/8$, where $n$ denotes the number of channels in the corresponding block). \textbf{(b)} Vision transformers (ViT). Top, schematic of a transformer block and multi-head self-attention with low-rank feedback channels; low rank feedback used to train all weights, including MLP and attention weights. Bottom, training curves (test accuracy versus training step) for BP (black) and low-dimensional feedback with a shared feedback rank $r$ across all linear maps (color-coded); inset shows the corresponding training loss. \textbf{(c)} Scaling and efficiency in ViT. Top, holding the feedback rank fixed ($r=36$) while reducing the embedding dimension decreases accuracy, indicating that performance continues to benefit from larger feedforward representations even under a fixed low-dimensional error channel. Bottom, ViT test accuracy (bars) and total compute required to reach $90\%$ of the final accuracy (line, TFLOPs) as a function of feedback rank, revealing an intermediate-rank regime that preserves BP-level accuracy while reducing end-to-end training compute (about $20\%$ in these experiments).}
    \label{fig:nonlinear}
\end{figure}

\paragraph{Scaling with network width.}
Constraining the feedback to be low-rank changes the credit-assignment channel but leaves the feedforward model class unchanged. The representational capacity remains set by the width and depth of the forward network, not by the rank of the feedback matrices. Consistent with this, increasing the size of the feedforward network improves accuracy even when the feedback rank is held fixed (\cref{fig:nonlinear}c top). At the same time, our linear analysis above indicate that the rank required to match backpropagation accuracy is controlled primarily by task dimensionality $d$ (e.g., performance degrades when $r<d$). Together, these observations imply that one can benefit from overparameterized feedforward representations while keeping the feedback dimensionality tied to task complexity rather than to network width.

\paragraph{Computational efficiency.}
A direct benefit of low-rank feedback is a reduced cost of propagating errors through depth. For a layer with widths $n_l$ and $n_{l+1}$, backpropagation multiplies by $W_l^\top$ at cost $O(n_ln_{l+1})$ per example (or per token). With low-rank feedback $B_l=Q_lP_l$ of rank $r$, the same operation factors into $P_l\boldsymbol{\delta}{l+1}$ followed by $Q_l(\cdot)$, at cost $O(r(n_l+n_{l+1}))$. Accordingly, the FLOPs associated with backward error propagation decrease linearly with the effective error dimensionality $r$.

Total training cost reflects three additional effects. First, the factors $Q_l$ and $P_l$ must be updated, but these updates depend only on $W_l$ and do not scale with batch size, so they can be amortized by updating the feedback factors at longer intervals. Second, extremely small ranks can slow convergence; in the regime where final performance matches backpropagation, learning curves are comparable (\cref{fig:nonlinear}c top), whereas at very low ranks, additional epochs may be required. Third, the backward pass is only part of training, which also includes the full-rank forward pass and the weight updates.

To capture these effects, we measure the total number of FLOPs required to reach $90\%$ of the final test accuracy, jointly accounting for backward-pass savings, feedback-factor updates, and any change in convergence time. This yields a U-shaped dependence on rank: at large $r$, per-iteration cost approaches backpropagation, whereas at very small $r$ slower convergence dominates. For a vision transformer on CIFAR-10, an intermediate rank achieves the best trade-off, reducing total compute (by about $20\%$ in our experiments) while maintaining backpropagation-level accuracy (\cref{fig:nonlinear}c bottom).

\section{Local learning rules}
\label{sec:local_rule}
The factorized feedback-learning scheme above achieves backpropagation-level performance with an error channel of rank $r \ll n_l$, and yields tangible compute savings. However, at its core it remains a surrogate for weight transport since the update of $(Q_l,P_l)$ is driven by the corresponding forward matrix $W_l$ through the objective $\|Q_lP_l - W_l^\top\|_F^2$. Implementing such ``learning through distance'' would require synapses in the feedback pathway to access parameters in a separate, remote pathway, which is difficult to reconcile with known biological plasticity mechanisms. Moreover, explicitly approximating $W_l^\top$ presupposes a parallel forward route whose weights are being transposed; this assumption breaks down when error signals arrive through indirect pathways that do not mirror the feedforward circuit, as in direct feedback alignment~\cite{nokland2016direct}. These limitations motivate a complementary perspective: rather than learning a proxy for $W_l^\top$, we seek a genuinely local mechanism that learns a useful low-dimensional error representation while keeping the remaining feedback circuitry fixed.

Our linear analysis suggests that, under a rank constraint, the crucial degree of freedom is not an element-wise approximation of $W_l^\top$ but the \emph{error subspace} that the feedback pathway can represent and transmit. Writing $B_l = Q_l P_l$, the matrix $P_l \in \mathbb{R}^{r\times m}$ projects output-space errors into an $r$-dimensional channel, so the columns of $P_l^\top$ span the subspace of output error directions that can be assembled upstream. When this subspace is poorly aligned with the task, the stationarity condition becomes underdetermined and training can converge to spurious fixed points. Conversely, when $P_l^\top$ spans task-relevant directions, the feedback pathway preserves informative components of the error while discarding directions that are redundant for learning. 

Ideally, $P_l$ has $r$ orthonormal rows ($P_l P_l^\top = I_r$), so that $P_l^\top P_l$ is the rank-$r$ projector onto the transmitted subspace in the downstream error space. Thus, $P_l$ learns the top-$r$ subspace of the incoming teaching signal to layer $l$ (i.e., the vector that $P_l$ multiplies). Concretely, we want the transmitted subspace to align with the top principal components of the layerwise error, given by the eigenvectors of $\frac{1}{p}\sum_\mu \boldsymbol{\delta}_{l+1}^\mu \boldsymbol{\delta}_{l+1}^{\mu\top} $, where $\boldsymbol{\delta}_{l+1}^\mu$ denotes the locally available teaching/error signal layer $l+1$ for example $\mu$. Specifically, at the output layer, the error is given by $\boldsymbol{\delta}_L^\mu=\partial \mathcal{L}^\mu/\partial \hat{\boldsymbol y}^\mu$, while for hidden layers it is the error signal broadcast through the constrained feedback pathway. This choice allocates the limited rank of the feedback channel to the directions along which the network is currently making the biggest mistakes, and it uses a signal that is locally available wherever feedback arrives.

To learn the principal components of the error using online updates, we use a multi-component Oja rule driven by the locally available teaching signal~\cite{oja1982simplified},
\begin{equation}\label{eq:DeltaP}
    \Delta P_l^\mu
    = \eta\, P_l\, \boldsymbol{\delta}_{l+1}^\mu \boldsymbol{\delta}_{l+1}^{\mu\top}\, (I - P_l^\top P_l) - \lambda P_l,
\end{equation}
where $I$ is the identity matrix. 

Eq.~\eqref{eq:DeltaP} is a matrix-valued Oja update for online principal-subspace learning. A Hebbian term driven by the teaching signal $\boldsymbol{\delta}_{l+1}^\mu$ increases the components of $P_l$ along directions of high error variance, while the remaining factors stabilize the dynamics and set the overall scale. In particular, the factor $(I-P_l^\top P_l)$ promotes row-orthonormality of $P_l$, and the decay term $-\lambda P_l$ controls the magnitude of the weights.

Although the orthonormalizing factor is written compactly as a matrix product, its role is to express within-population competition and normalization in the $r$-dimensional error channel. In Oja's original single-component rule, the stabilizing term can be interpreted as a postsynaptic-activity--dependent normalization of an otherwise Hebbian increment, yielding an online update that converges to the leading principal component~\cite{oja1982simplified}. Extending PCA learning to multiple components necessarily requires a mechanism that prevents different components from collapsing onto the same direction, which is achieved by introducing competition among the units representing the learned subspace.

Several biologically motivated constructions realize this competition using only local interactions within the postsynaptic population~\cite{sanger1989optimal, jankovic2003new, jankovic2006modulated, minden2018biologically, isomura2018error, pehlevan2015hebbian}. In particular, Sanger's generalized Hebbian algorithm implements multi-component PCA by combining Hebbian plasticity with within-layer interactions that subtract already-learned projections, thereby enforcing orthogonality through local competition among units~\cite{sanger1989optimal}. More generally, similarity-matching formulations derive networks in which feedforward synapses learn Hebbianly while lateral synapses learn anti-Hebbianly, and the lateral dynamics locally decorrelate and normalize neural activities, effectively playing the role of the orthonormalization implicit in \eqref{eq:DeltaP}~\cite{pehlevan2015hebbian,pehlevan2017similarity}. In this sense, \eqref{eq:DeltaP} should be read as a compact description of a circuit-level computation in which synaptic updates depend on pre- and postsynaptic activity, while orthogonality and normalization are maintained by local competition within the error-channel population.

Finally, Because $P_l$ adapts gradually, it tracks the dominant modes of the nonstationary teaching signal; the forward weights then reduce those modes, causing the subspace to rotate as learning progresses.

Together with the layer-wise weight update \eqref{eq:Wlupdate}, this yields a local learning scheme in which feedback synapses update from signals available at their pre- and postsynaptic terminals, and stabilization is mediated by within-layer normalization and competition rather than by nonlocal parameter access. 

We refer to this scheme as \emph{low-dimensional feedback alignment} (LDFA). In what follows, we show that learning $P_l$ alone, with $Q_l$ fixed and random, is sufficient to reach backpropagation-level accuracy; as in Feedback Alignment~\cite{lillicrap2016random}, the forward weights can then align to the fixed mixing induced by $Q_l$.

\subsection{Linear analysis}
We analyze the coupled learning dynamics in the one-hidden-layer linear network of Section~\ref{sec:linear}, using the same rotation into the singular basis of the input-output correlation $\Sigma_{io}=USV^\top$. Writing $W_1=\bar W_1 V^\top$, $W_2=U\bar W_2$, and $P=\bar P U^\top$ yields an effective feedback $B=QP=Q\bar P U^\top$, with $Q$ fixed and $\bar P(t)$ learned. In the continuous-time limit, the forward dynamics in the rotated basis are exactly those of \cref{eq:Wdynamics}, with the sole modification that $\bar B(t)=Q\bar P(t)$ is now time-dependent.

The new ingredient is the evolution of $\bar P$ under the error-driven Oja rule \eqref{eq:DeltaP}. For squared loss and whitened inputs, the output-layer error in the rotated coordinates satisfies $\bar{\boldsymbol{\delta}}^\mu(t)=\bar E(t)\,\bar{\boldsymbol x}^\mu$, where $\bar{\boldsymbol x}^\mu=V^\top \boldsymbol x^\mu$ and $\bar E(t)=S-\bar W_2(t)\bar W_1(t)$ is the residual in the singular basis. Averaging over samples, therefore, gives an error covariance $\bar E(t)\bar E(t)^\top$, and the continuous Oja update can be written as
\begin{equation}\label{eq:Pdynamics}
    \tau \frac{d\bar{P}}{dt}
    = \bar{P}\,\bar E\bar E^\top\,(I-\bar{P}^\top \bar{P})-\lambda \bar{P}.
\end{equation}
Equations~\eqref{eq:Wdynamics} and \eqref{eq:Pdynamics} thus define a coupled first-order dynamical system for $(\bar W_1,\bar W_2,\bar P)$.

\begin{figure}
    \centering
    \includegraphics[width=1.0 \linewidth]{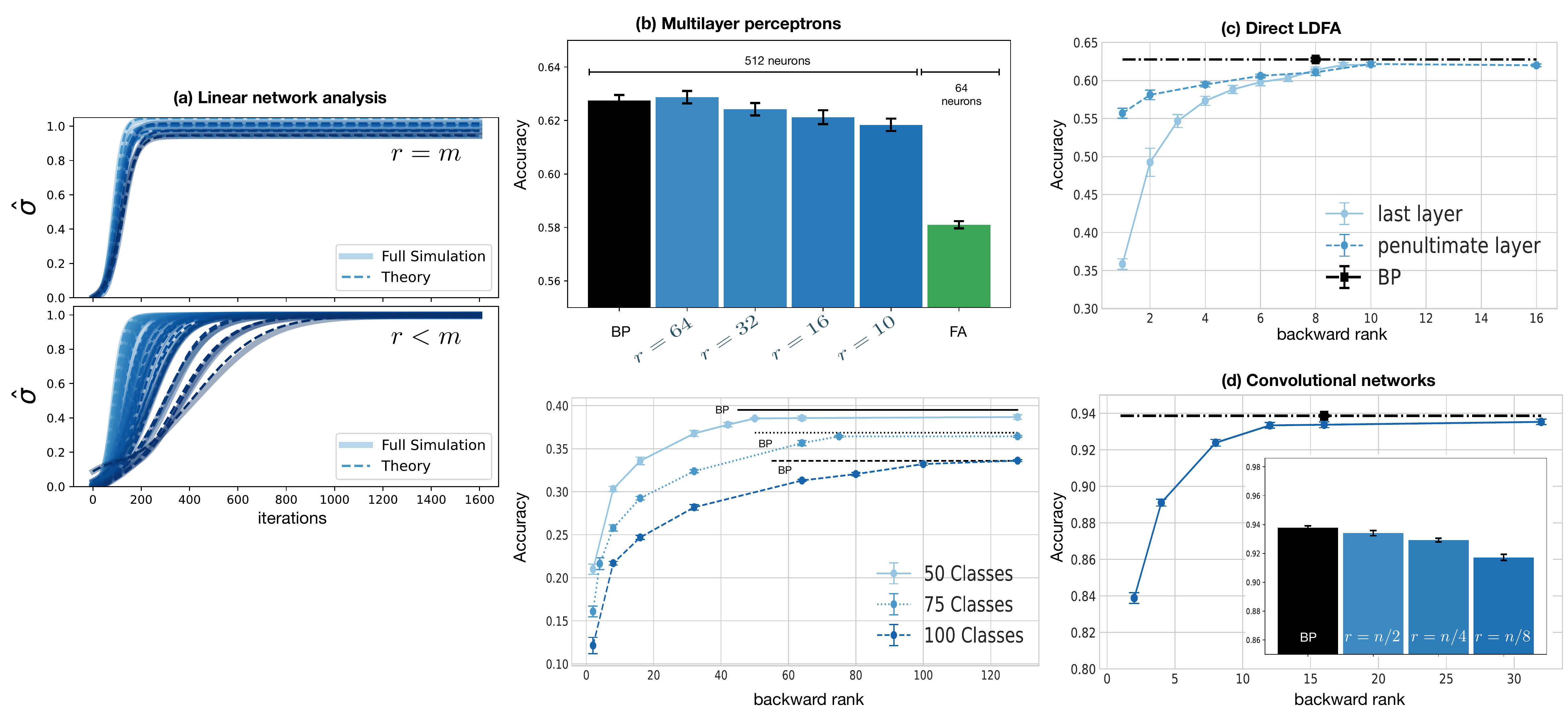}
\caption{\textit{Low-dimensional feedback alignment (LDFA) as a synaptically local error pathway.} \textbf{(a)} Linear-network analysis ($n=128$, $k=64$, $m=64$) with factorized feedback $B=QP$. Solid curves show simulations and dashed curves show the analytical prediction for the evolution of mode-wise singular values. With full-rank feedback ($r=m$), all task modes are recovered, whereas with rank-constrained feedback ($r<m$) learning prioritizes the dominant task modes selected by the Oja dynamics. \textbf{(b)} Multilayer perceptrons on CIFAR. Top: separating feedforward capacity from feedback dimensionality. A wide MLP reaches high accuracy even when the teaching signal is strongly compressed (LDFA with low rank $r$), whereas a much narrower MLP performs substantially worse despite having an unconstrained (full-rank) error pathway (BP). Bottom: varying task output dimensionality by subsampling classes from CIFAR-100 shows that the minimal rank required to match BP scales primarily with task dimension (number of classes), not with hidden-layer width (all layers use the same rank $r$).
\textbf{(c)} Direct LDFA implements direct error projection by broadcasting a low-dimensional teaching signal to each layer either from the output layer (solid) or the penultimate layer (dashed), approaching BP-level accuracy as rank increases. \textbf{(d)} Convolutional networks trained with LDFA. A 4-block VGG-style CNN attains BP-level performance with low backward rank even when the widest blocks contain 512 channels. Inset: constraining all layers to a fixed fraction of their width ($r=n/2,n/4,n/8$) yields graceful degradation.}

    \label{fig:LDFA}
\end{figure}

This system has a straightforward interpretation. The Oja flow \eqref{eq:Pdynamics} steers the transmitted subspace toward the leading eigenvectors of the current error covariance, so the rank-$r$ channel allocates its constrained capacity to the modes that dominate the residual. As learning progresses, this mechanism tracks the dominant modes of the current residual, which in this solvable setting biases learning toward the leading singular directions of $\Sigma_{io}$ (Figure \ref{fig:LDFA}a).

\subsection{Deep nonlinear networks}
We next test whether the local rule extends beyond the linear setting to deep nonlinear networks. Following the same procedure in \cref{sec:deep_nonlinear}, we train MLPs on CIFAR using the same constrained feedback architecture as above, where each layer receives an $r$-dimensional error signal through a fixed random mixing $Q_l$, while the projection $P_l$ is learned locally by the error-driven Oja update \eqref{eq:DeltaP}. Across architectures, we find that as with the normative approach in \cref{sec:linear}, rank required to match backpropagation tracks \emph{task dimensionality} rather than network size. Concretely, when we subsample CIFAR-100 to classification problems with 50, 65 or 100 classes and constrain all feedback matrices to rank $r$, performance approaches full backpropagation once $r$ reaches the output dimension defined by the number of classes (Fig.~\ref{fig:LDFA}b).

\subsection{Low-dimensional error broadcasting}
A key limitation of the normative feedback-learning rule in \cref{sec:linear} is its explicit dependence on the forward weights $W_l$, in the minimization of  \cref{eq:BackwardLoss}. This effectively assumes reciprocal connections between consecutive layers, and therefore cannot be applied to error pathways that bypass intermediate areas. In biological circuits, however, instructive signals are known to broadcast through indirect and long-range projections, while a mechanism that rely on  local reciprocals has yet been found. A canonical example is the dopaminergic reward-prediction-error signal, which is delivered to many downstream targets via widespread projections~\cite{Montague1996Framework,Schultz1997Neural,Steinberg2013Causal}.

Direct feedback alignment (DFA) was introduced precisely to relax the requirement of reciprocal connectivity by broadcasting the output error directly to each hidden layer through fixed random feedback matrices~\cite{nokland2016direct}. In its simplest form, each layer receives a teaching signal computed in parallel from the output-layer error, enabling updates without sequential backward transport through intermediate layers. While DFA can train deep networks in some regimes, its naive formulation typically exhibits a substantial performance gap relative to backpropagation on modern vision benchmarks, and convolutional architectures are a particularly challenging case~\cite{bartunov2018assessing, crafton2019direct}.

Our local rule removes the architectural constraint that prevented the normative approach from operating in this regime. Concretely, for each layer $l$ we broadcast the error from a chosen source layer $l'$ (often $L$) directly to layer $l$ using a low-rank matrix,
\begin{equation}\label{eq:DirectLDFA}
\boldsymbol{\delta}_l = \big(Q_{ll'} P_{ll'}\boldsymbol{\delta}_{l'}\big)\odot f'(a_l),\quad l<l'\leq L
\end{equation}
where $l'$ is an efferent layer to $l$ or the output error $\delta_L$. Same as above, $Q_{ll'}$ is fixed and random and $P_{ll'}\in\mathbb{R}^{r\times n_{l'}}$ is learned by the local Oja update \cref{eq:DeltaP} (with $\boldsymbol{\delta}_{l'}$ as its driving signal). This construction preserves the defining feature of DFA, namely parallel error broadcast through non-reciprocal pathways, while endowing the broadcast channel with an adaptive, low-rank projection that allocates its limited capacity to the dominant directions of the task error. Because the update of $P_{ll'}$ is unsupervised and depends only on the locally available teaching signal, it applies unchanged when the feedback edges skip layers or originate from alternative sources (e.g., the penultimate layer), and it does not require access to the forward weights.

Empirically, this direct error projection closes much of the gap between DFA and backpropagation in deep nonlinear networks. In particular, learning $P_l$ with $Q_l$ fixed suffices to train deep MLPs and CNNs with direct error broadcasts to backpropagation-level performance (Fig.~\ref{fig:LDFA}c), including regimes in which standard DFA is known to struggle. This demonstrates that the local subspace-learning mechanism is compatible with non-mirrored, indirect error pathways, strengthening the biological motivation of the framework while retaining the benefits of low-dimensional feedback.

\subsection{Convolutional networks}
A particularly stringent test is the convolutional setting. In deep CNNs, naive alignment schemes often exhibit a substantial gap relative to backpropagation unless augmented with additional mechanisms that improve alignment or stabilize training. Our low-dimensional feedback framework remains effective in this regime: we train CNNs reliably and reach backpropagation-level accuracy at comparable rank (Fig.~\ref{fig:nonlinear}b), despite the combined challenges of deep credit assignment, weight sharing, and a restricted error channel.

In all CNN experiments we learn the error projection $P_l$ using the local Oja update \eqref{eq:DeltaP}, which is the core mechanism that allocates the rank-$r$ channel to task-relevant error directions. Convolutional architectures, however, are a particularly stringent regime for alignment-based learning, and prior work on feedback alignment has typically found that allowing some adaptation of the feedback weights substantially improves robustness and accuracy in deep CNNs~\cite{kolen1994backpropagation, akrout2019deep}. Following this established practice, we therefore also allow the mixing matrix $Q_l$ to adapt via a simple synaptically local Hebbian rule,
\begin{equation}\label{eq:DeltaQ}
\Delta Q_l^\mu = \eta\, \boldsymbol{h}_l^\mu \big(P_l \boldsymbol{\delta}_{l+1}^\mu\big)^\top - \lambda Q_l,
\end{equation}
which preserves the low-rank structure and does not introduce weight transport or any dependence on forward parameters. Importantly, \eqref{eq:DeltaQ} remains local: the presynaptic activity $\boldsymbol{h}_l^\mu$ and the postsynaptic teaching signal $P_l\boldsymbol{\delta}_{l+1}^\mu$ are available at the feedback synapse, and the update depends only on these signals and the synaptic weight. Empirically, this augmentation is consistent with the FA literature on challenging vision models and improves performance in convolutional architectures while retaining the central feature of LDFA: learning a low-dimensional error subspace through $P_l$.

Figure \ref{fig:LDFA}d shows that training CNNs with low-dimensional error and local-only learning rules can match backpropagation performance, while maintaining more biological integrity.

\section{Error dimensionality shapes neural receptive fields}
\label{sec:ReceptiveFields}
\begin{figure}
    \centering
    \includegraphics[width=1.0\linewidth]{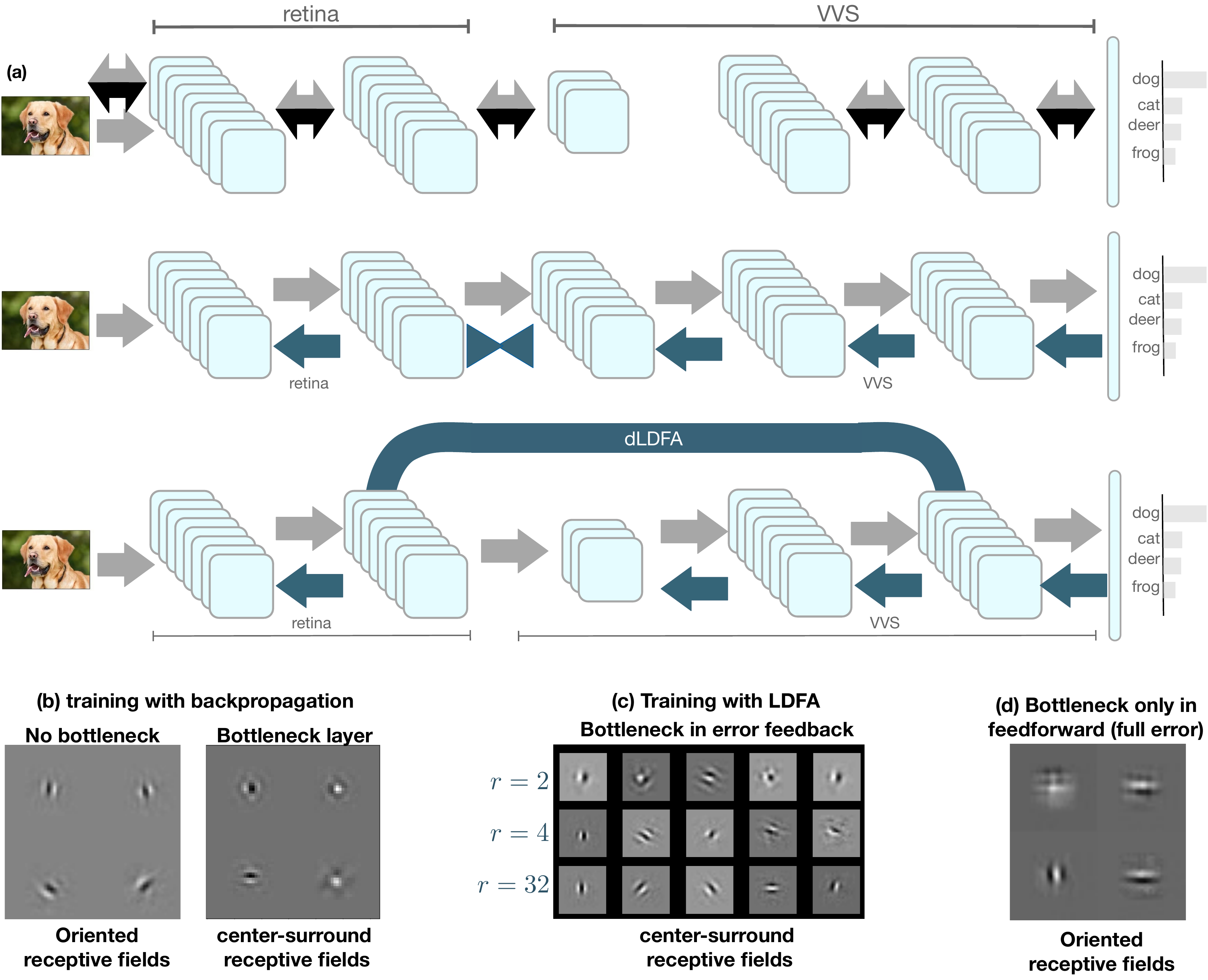}
\caption{\textit{Receptive fields are shaped by feedback dimensionality.} \textbf{(a)} Simplified model of the early ventral visual stream (retina $\rightarrow$ VVS) adapted from~\cite{lindsey2019unified}, and two manipulations that decouple feedforward bottlenecks from the teaching pathway. Top: an anatomical bottleneck limits the retinal output. Middle: the feedforward architecture is left full-width, but the teaching signal to the retinal layer is restricted by a low-rank feedback map (LDFA). Bottom: the feedforward bottleneck is reinstated, while the retinal layer receives a high-dimensional teaching signal via direct error projection (dLDFA; Eq.~(16)), bypassing backward compression. \textbf{(b)} Backpropagation reproduces the Lindsey et al.\ phenomenology: removing the retinal bottleneck yields oriented, edge-like receptive fields, whereas a narrow retinal bottleneck yields center-surround receptive fields. \textbf{(c)} With the feedforward pathway unconstrained but feedback to the retina rank-limited, LDFA reliably produces center-surround receptive fields; decreasing the feedback rank $r$ (examples $r=32,4,2$) increases their effective rotational symmetry. \textbf{(d)} Conversely, with a feedforward retinal bottleneck but high-dimensional error delivered directly to the retina using direct LDFA (dLDFA), receptive fields become oriented. Together, these manipulations show that the dimensionality and routing of the teaching signal can reproduce, and even override, receptive-field structure previously attributed to feedforward anatomical constraints.}

    \label{fig:conv}
\end{figure}

At the heart of our framework is the ability to decouple the error-feedback pathway from the feedforward pass, and to control the feedback channels independently. In Secs.~\ref{sec:linear} and \ref{sec:deep_nonlinear} we used this decoupling to study how restricting the backward pathway controls the efficiency of learning, and in Sec.~\ref{sec:local_rule} we derived local update rules (including convolutional networks) that implement these constraints using synapse-local information. Here we ask a complementary question: beyond efficiency, does the structure of the error pathway shape the representations learned in early layers? Concretely, we study how receptive fields of shallow layers change when we constrain the dimensionality of the error signal reaching them.

The relation between architecture and learned receptive fields has been studied before. Lindsey et al.\ showed that in anatomically constrained convolutional models, narrowing the retinal output bottleneck yields center-surround receptive fields in the retinal layer, whereas widening it yields orientation-selective receptive fields~\cite{lindsey2019unified}. We hypothesize that this transition is driven primarily by the dimensionality of the teaching signal available to the retinal layer. With constrained feedback channels, the compressed error signal  biases learning toward higher-symmetry solutions, even when the feedforward pathway is left unconstrained.

LDFA provides a direct way to test this hypothesis because it lets us constrain the backward pathway while keeping the forward architecture fixed, and, via direct error projection (Eq.~\eqref{eq:DirectLDFA}), to manipulate error delivery independently of feedforward bottlenecks. We first remove the feedforward bottleneck by using full-width layers throughout, and restrict only the backward pathway by training a rank-$r$ feedback matrix to the retinal layer using LDFA (Fig.~\ref{fig:conv}a). We then visualize the resulting retinal receptive fields using standard feature-visualization methods~\cite{erhan2009visualizing} (Fig.~\ref{fig:conv}b,c). Despite the unconstrained feedforward architecture, restricting the feedback rank reliably produces center-surround receptive fields, and decreasing $r$ increases their effective rotational symmetry.

We next perform the complementary manipulation. We reinstate a narrow feedforward bottleneck as in~\cite{lindsey2019unified}, but ensure that the retinal layer receives a high-dimensional error signal by using direct error projections (dLDFA; Eq.~\eqref{eq:DirectLDFA}), thereby bypassing any backward compression (Fig.~\ref{fig:conv}a bottom). In this setting, the retinal receptive fields become orientation selective (Fig.~\ref{fig:conv}d), matching the wide-bottleneck regime of~\cite{lindsey2019unified}. Together, these experiments show that the feedback pathway can reproduce, and even override, receptive-field phenomenology previously attributed to feedforward anatomical constraints and that error dimensionality acts as a powerful inductive bias on early-layer selectivity.

The Lindsey et al.\ architecture is intentionally simplified and should not be read as a mechanistic model of retinal development; in particular, there is no evidence for supervised error signals propagating from cortex or thalamus back to the retina, and early retinal representations are widely thought to be shaped largely by local and unsupervised mechanisms~\cite{zhuang2021unsupervised}. Our claim is therefore methodological and conceptual. Within a framework in which forward and backward passes are decoupled and implementable with local rules, the dimensionality and routing of learning signals can act as an inductive bias on receptive-field structure. This perspective applies most directly to early sensory stages that plausibly receive low-dimensional instructive signals (e.g., via neuromodulation or top-down pathways), and suggests that constraints on the error pathway may shape emergent selectivity at least as strongly as constraints on the feedforward circuit itself. This complements goal-driven modeling of sensory cortex by highlighting the backward pathway as an additional axis of inductive bias, alongside architecture and objective~\cite{yamins2016using}.

\section{Discussion}
Our central result is that deep networks do not require transmitting the full layerwise gradient: it is sufficient to deliver a low-dimensional teaching signal to each layer, provided the error is projected into task-relevant directions. In a solvable linear theory, fixed low-rank feedback generically makes the stationarity conditions underdetermined and introduces spurious fixed points. Learning the feedback factors aligns the transmitted subspace with the evolving error and restores backpropagation-like convergence across nonlinear networks, including CNNs and vision transformers.

This perspective cleanly separates a normative question (which error subspace should be communicated) from an implementational one (how that subspace can be learned). We first derive feedback learning as a low-rank factorization objective whose updates do not scale with batch size, and then show how essentially the same subspace can be learned with synapse-local plasticity driven by the arriving error, enabling direct error projection through non-mirrored pathways and revealing error dimensionality as an inductive bias that can shape early receptive fields.

\paragraph{Computational efficiency.} 
Because low-rank feedback constrains the error channel itself, the expensive multiplication by $W_l^\top$ in the backward pass is replaced by two thin multiplies, reducing per-example cost from $O(n_ln_{l+1})$ to $O(r(n_l+n_{l+1}))$ and making backward FLOPs scale linearly with the effective error dimensionality $r$. End-to-end, this creates a trade-off in which decreasing the rank $r$ lowers computational cost, but excessive compression degrades learning efficiency, requiring more samples. Consequently, total compute exhibits a U-shaped dependence on $r$, with an intermediate rank minimizing FLOPs-to-accuracy in our transformer experiments.  This positions low-dimensional feedback as a structured alternative to gradient sparsification and compression methods such as \textit{meProp} \cite{sun2017meprop} and \textit{Deep Gradient Compression} \cite{lin2018deepgradientcompression}. Rather than sparsifying a full gradient after it is formed, we learn the projection that determines which error directions are represented upstream, so the savings come from never materializing or transporting the discarded components.

\paragraph{Aligning to the error space.}
At first glance, our local rule resembles adaptive variants of Feedback Alignment (FA), where the feedback weights are trained to better match the feedforward weights and thereby approximate backpropagation~\cite{lillicrap2016random,akrout2019deep}. The factorization $B_l=Q_lP_l$ reveals a different target: $Q_l$ only mixes an $r$-dimensional teaching channel into layer $l$, whereas $P_l$ determines which output-error directions can be represented and transmitted upstream (the subspace spanned by $P_l^\top$). Learning $P_l$ therefore \textbf{aligns the transmitted error subspace with the dominant modes of the current residual}, improving the information content of the teaching signal itself. This perspective explains why fixed low-rank feedback can induce spurious fixed points, and why subspace learning resolves them: the bottleneck fails when it discards task-relevant error directions, not merely when it misaligns with $W_l^\top$.

\paragraph{Low-rank adaptation (LoRA) in large language models.} 
Recent parameter-efficient fine-tuning methods for large language models, most notably LoRA, achieve strong adaptation by freezing pretrained weights and learning low-rank updates $\Delta W$ within transformer layers \cite{hu2022lora}, with practical extensions that push memory efficiency further (e.g., quantization with QLoRA) and adaptive rank allocation)~\cite{dettmers2023qlora,liu2024dora}. Our approach is complementary in where the low-rank constraint is imposed. Here, instead of restricting the forward update parameterization while retaining full backpropagated errors, we restrict and learn the \emph{error subspace} communicated to each layer. This provides a middle ground in which the number of trainable forward parameters can grow (increasing representational capacity) while the cost of credit assignment remains controlled by the feedback rank. 

\paragraph{Bridging biological constraints and gradient-based learning.}
Our work addresses a basic anatomical mismatch between backpropagation and cortex: feedback pathways are not organized as a synapse-wise transpose of the forward weights and need not deliver a neuron-by-neuron teaching signal~\cite{markov2014anatomy}. Predictive coding and equilibrium propagation propose mechanistic routes by which local plasticity, together with inference dynamics, can approximate gradient-based learning in hierarchical circuits~\cite{rao1999predictive,whittington2017approximation,scellier2017equilibrium}. Our contribution is complementary: we remain in the standard objective-minimization setting, but treat the teaching pathway as a constrained resource and learn a low-dimensional projection of error signals that preserves task-relevant directions. This yields effective credit assignment under severe bandwidth limits and, via direct error projection, remains compatible with indirect and non-reciprocal delivery of instructive signals.

\paragraph{Low-dimensional teaching signals and reward pathways.}
The same perspective offers a middle ground between scalar reinforcement signals and full gradient transport. Midbrain dopamine has long been linked to reward-prediction-error teaching signals, yet recordings increasingly suggest heterogeneous coding in which dopaminergic populations and terminals reflect multiple sensory, motor, and cognitive variables beyond a single scalar reward~\cite{cohen2012neuron,parker2016reward,engelhard2019specialized}. In our framework, such signals can be interpreted as a low-dimensional teaching vector: a compressed, broadcast error representation whose efficacy depends on learning the appropriate projection into each layer’s plastic synapses. This suggests a concrete computational role for low-dimensional neuromodulatory feedback, without requiring the brain to propagate a full high-dimensional gradient through deep circuits.

\paragraph{Feedback structure as an inductive bias.}
Inductive biases in deep learning are usually discussed as properties of the forward architecture, where weight sharing, locality, and symmetry constraints bias learning toward particular solution classes and improve generalization~\cite{zhou2020gnn_review,bronstein2017geometric,soudry2018implicit_bias,goyal2022inductive_biases}. Our results suggest that the \emph{teaching channel} can play a similar role by constraining and shaping the feedback pathway. The feedback path controls which error directions are expressible upstream, thereby biasing optimization toward representations that are easiest to learn under the available error bandwidth. This viewpoint is consistent with recent analyses showing that feedback-alignment-style learning exhibits implicit regularization and can select among degenerate low-loss solutions~\cite{refinetti2021align}. In our setting, the receptive-field experiments provide a concrete illustration that modifying only the feedback pathway can systematically alter learned symmetries. A promising direction is therefore to treat feedback design as a new knob for \emph{algorithmic} inductive bias: for example, imposing topographic, convolutional, modular, or sign-constrained structure on low-rank feedback may preferentially promote invariances, robustness, or task-specific feature reuse, while maintaining computational efficiency through a controlled error bottleneck.

In summary, we argue that the essential constraint for scalable credit assignment is not access to a full, layerwise gradient but access to a \emph{learned} low-dimensional teaching channel that preserves task-relevant error directions. This reframes feedback as a resource that can be optimized, offering a principled route to more efficient training when backward computation or communication is limiting, and a concrete hypothesis for how anatomically restricted pathways could still support deep learning. More broadly, our results invite future work that treats the structure of feedback, not only the forward architecture, as a design axis that shapes both efficiency and the inductive biases of learned representations.

\section*{Acknowledgments}
This work was supported by the Gatsby Charitable Foundation.


\bibliography{references}


\renewcommand\theequation{\Alph{section}\arabic{equation}} 
\counterwithin*{equation}{section} 
\renewcommand\thefigure{\Alph{section}\arabic{figure}} 
\counterwithin*{figure}{section} 
\renewcommand\thetable{\Alph{section}\arabic{table}} 
\counterwithin*{table}{section} 

\newpage
\begin{appendices}
\section*{Appendix}

\section{Full linear theory}
In this section, we provide the analysis of the linear networks for both Feedback Alignment and LDFA.

\subsection*{Detailed analysis of Feedback Alignment learning dynamics}
From eq. (4) in the main text, the weight updates for a single sample \( \mu \) are given by:
\begin{equation}
\begin{aligned}
    \Delta W_1^\mu &= \eta B \left( \boldsymbol{y}^\mu - W_2 W_1 \boldsymbol{x}^\mu \right) \boldsymbol{x}^{\mu \top}, \\
    \Delta W_2^\mu &= \eta \left( \boldsymbol{y}^\mu - W_2 W_1 \boldsymbol{x}^\mu \right) \boldsymbol{x}^{\mu \top} W_1^\top,
\end{aligned}
\label{eq:DeltaW_mu}
\end{equation}
where:
\begin{itemize}
    \item \( \eta \) is the learning rate,
    \item \( \boldsymbol{x}^\mu \in \mathbb{R}^n \) is the input vector for sample \( \mu \),
    \item \( \boldsymbol{y}^\mu \in \mathbb{R}^m \) is the corresponding target output,
    \item \( W_1 \in \mathbb{R}^{k \times n} \) and \( W_2 \in \mathbb{R}^{m \times k} \) are the weight matrices,
    \item \( B \in \mathbb{R}^{k \times m} \) is a predefined matrix (e.g., a feedback or scaling matrix).
\end{itemize}

We introduce the empirical covariance matrices:
\begin{align}
    \Sigma_{io} &= \frac{1}{p} \sum_{\mu=1}^{p} \boldsymbol{y}^\mu \boldsymbol{x}^{\mu \top}, \label{eq:Sigma_io} \\
    \Sigma_{oo} &= \frac{1}{p} \sum_{\mu=1}^{p} \boldsymbol{y}^\mu \boldsymbol{y}^{\mu \top}, \label{eq:Sigma_oo} \\
    \Sigma_{ii} &= \frac{1}{p} \sum_{\mu=1}^{p} \boldsymbol{x}^\mu \boldsymbol{x}^{\mu \top} = I, \label{eq:Sigma_ii}
\end{align}
where \( \Sigma_{ii} = I \) assumes that the input vectors are whitened (i.e., have unit covariance). Summing over all \( p \) training examples, we obtain the average weight updates:
\begin{equation}
\begin{aligned}
    \Delta W_1 &= \frac{\eta}{p} \sum_{\mu=1}^{p} \Delta W_1^\mu \\
    &= \frac{\eta}{p} \sum_{\mu=1}^{p} B \left( \boldsymbol{y}^\mu - W_2 W_1 \boldsymbol{x}^\mu \right) \boldsymbol{x}^{\mu \top}.
\end{aligned}
\end{equation}

Using these definitions, the update for \( W_1 \) simplifies to:
\begin{align}
    \Delta W_1 &= \eta B \left( \Sigma_{io} - W_2 W_1 \Sigma_{ii} \right) \\
    &= \eta B \left( \Sigma_{io} - W_2 W_1 \right). \label{eq:DeltaW1}
\end{align}

Under the limit as \( \eta \rightarrow 0 \) with \( \eta = \frac{dt}{\tau} \) (where \( \tau \) is a time constant), we transition from discrete updates to continuous-time dynamics:
\begin{equation}
    \tau \frac{d W_1}{dt} = B \left( \Sigma_{io} - W_2 W_1 \right),
    \label{eq:W1_dynamics}
\end{equation}
which matches the weight dynamics presented in eq. (5) of the main text.

Similarly, the update for \( W_2 \) becomes:
\begin{align}
    \Delta W_2 &= \frac{\eta}{p} \sum_{\mu=1}^{p} \Delta W_2^\mu \\
    &= \frac{\eta}{p} \sum_{\mu=1}^{p} \left( \boldsymbol{y}^\mu - W_2 W_1 \boldsymbol{x}^\mu \right) \boldsymbol{x}^{\mu \top} W_1^\top \\
    &= \eta \left( \Sigma_{io} - W_2 W_1\right) W_1^\top,
\end{align}
which simplifies under the same limit to:
\begin{equation}
    \tau \frac{d W_2}{dt} = \left( \Sigma_{io} - W_2 W_1\right) W_1^\top.
    \label{eq:W2_dynamics}
\end{equation}
Eq. \eqref{eq:W1_dynamics} and eq. \eqref{eq:W2_dynamics} describe the continuous-time dynamics of the weights \( W_1 \) and \( W_2 \) under the given learning rule.

We consider the singular value decomposition (SVD) of the covariance matrix \( \Sigma_{io} \):
\begin{equation}
    \Sigma_{io} = U S V^\top,
    \label{eq:SVD}
\end{equation}
where:
\begin{itemize}
    \item \( U \in \mathbb{R}^{m \times d} \) and \( V \in \mathbb{R}^{n \times d} \) are matrices with orthonormal columns,
    \item \( S \in \mathbb{R}^{d \times d} \) is a diagonal matrix containing the singular values,
    \item \( d \) is the rank of \( \Sigma_{io} \).
\end{itemize}

We perform a rotation of the weight matrices and \( B \) as follows:
\begin{equation}
\begin{aligned}
    W_1 &= \bar{W}_1 V^\top, \\
    W_2 &= U \bar{W}_2, \\
    B &= \bar{B} U^\top.
\end{aligned}
\label{eq:rotations}
\end{equation}

Substituting these into the previous weight dynamics, we have for \( W_1 \):
\begin{align}
    \tau \frac{d W_1}{dt} &= B \left( \Sigma_{io} - W_2 W_1 \right) \\
    &= \bar{B} U^\top \left( U S V^\top - U \bar{W}_2 \bar{W}_1 V^\top \right).
\end{align}

Since \( U^\top U = I \) (due to orthonormal columns of \( U \)), we can simplify:
\begin{align}
    \tau \frac{d W_1}{dt} &= \bar{B} \left( U^\top U \right) \left( S - \bar{W}_2 \bar{W}_1 \right) V^\top \\
    &= \bar{B} \left( S - \bar{W}_2 \bar{W}_1 \right) V^\top.
\end{align}

Recognizing that \( W_1 = \bar{W}_1 V^\top \), we can write \( \frac{d W_1}{dt} = \frac{d \bar{W}_1}{dt} V^\top \). Thus, multiplying both sides on the right by \( V \) (since \( V^\top V = I \)):
\begin{equation}
    \tau \frac{d \bar{W}_1}{dt} = \bar{B} \left( S - \bar{W}_2 \bar{W}_1 \right).
    \label{eq:W1_rotated_dynamics}
\end{equation}

Similarly, for \( W_2 \):
\begin{align}
    \tau \frac{d W_2}{dt} &= \left( \Sigma_{io} - W_2 W_1\right) W_1^\top \\
    &= \left( USV\top - U \bar{W}_2 \bar{W}_1 V^\top \right) \left( \bar{W}_1 V^\top \right)^\top \\
\end{align}

Using the fact that \( V^\top V = I \) we simplify:
\begin{align}
    \tau \frac{d W_2}{dt} &= \left( U S - U \bar{W}_2 \bar{W}_1 \right) \bar{W}_1^\top \\
\end{align}

Since \( W_2 = U \bar{W}_2 \), we have \( \frac{d W_2}{dt} = U \frac{d \bar{W}_2}{dt} \). Multiplying both sides on the left by \( U^\top \):
\begin{equation}
    \tau \frac{d \bar{W}_2}{dt} = \left( S - \bar{W}_2 \bar{W}_1 \right) \bar{W}_1^\top.
    \label{eq:W2_rotated_dynamics}
\end{equation}

Eqs. \eqref{eq:W1_rotated_dynamics} and \eqref{eq:W2_rotated_dynamics} describe the dynamics of the rotated weights \( \bar{W}_1 \) and \( \bar{W}_2 \):
\begin{equation}
\begin{aligned}
    \tau \frac{d \bar{W}_1}{dt} &= \bar{B} \left( S - \bar{W}_2 \bar{W}_1 \right), \\
    \tau \frac{d \bar{W}_2}{dt} &= \left( S - \bar{W}_2 \bar{W}_1 \right) \bar{W}_1^\top.
\end{aligned}
\end{equation}

\subsection*{Derivation of Normative Low-Rank Feedback Learning}
Here we derive the learning dynamics for the normative variant of LDFA, where the feedback matrix \( B = QP \) is trained to explicitly approximate the transpose of the forward weights \( W_2^\top \). The feedback factors \( Q \in \mathbb{R}^{k \times r} \) and \( P \in \mathbb{R}^{r \times m} \) are updated to minimize the reconstruction loss:
\begin{equation}
    \mathcal{L}_B = \frac{1}{2} \| QP - W_2^\top \|_F^2.
\end{equation}
The gradients with respect to the feedback factors are:
\begin{equation}
\begin{aligned}
    \frac{\partial \mathcal{L}_B}{\partial P} &= Q^\top (QP - W_2^\top), \\
    \frac{\partial \mathcal{L}_B}{\partial Q} &= (QP - W_2^\top) P^\top.
\end{aligned}
\end{equation}
Assuming gradient descent with a time constant \( \tau_B \), the continuous-time dynamics are:
\begin{equation}
\begin{aligned}
    \tau_B \frac{dP}{dt} &= Q^\top (W_2^\top - QP), \\
    \tau_B \frac{dQ}{dt} &= (W_2^\top - QP) P^\top.
\end{aligned}
\label{eq:normative_dynamics_raw}
\end{equation}
We now rotate these dynamics into the singular basis of \( \Sigma_{io} \) using the same transformations as above: \( W_2 = U \bar{W}_2 \) (implying \( W_2^\top = \bar{W}_2^\top U^\top \)) and \( P = \bar{P} U^\top \). Note that \( Q \) operates in the hidden space and is not rotated by \( U \).

Substituting these into the dynamics for \( P \):
\begin{align}
    \tau_B \frac{d(\bar{P} U^\top)}{dt} &= Q^\top \left( \bar{W}_2^\top U^\top - Q \bar{P} U^\top \right) \\
    &= Q^\top \left( \bar{W}_2^\top - Q \bar{P} \right) U^\top.
\end{align}
Multiplying by \( U \) on the right yields the rotated dynamics for \( \bar{P} \):
\begin{equation}
    \tau_B \frac{d\bar{P}}{dt} = Q^\top \left( \bar{W}_2^\top - Q \bar{P} \right).
\end{equation}
Similarly for \( Q \):
\begin{align}
    \tau_B \frac{dQ}{dt} &= \left( \bar{W}_2^\top U^\top - Q \bar{P} U^\top \right) ( \bar{P} U^\top )^\top \\
    &= \left( \bar{W}_2^\top - Q \bar{P} \right) U^\top U \bar{P}^\top \\
    &= \left( \bar{W}_2^\top - Q \bar{P} \right) \bar{P}^\top.
\end{align}
Together with the forward weight updates derived in Eq. (A19) and (A25), this forms the closed dynamical system described in Eq. (10) of the main text.

\subsection*{Full derivation of local low-dimensional feedback alignment (LDFA) learning dynamics}
We consider an alternative algorithm where the matrix \( B \) is replaced by \( B = Q P \), with \( Q \in \mathbb{R}^{k \times r} \) and \( P \in \mathbb{R}^{r \times m} \).

The updates for \( Q \) and \( P \) for a single sample \( \mu \) are given by:
\begin{equation}
\begin{aligned}
    \Delta Q^\mu &= \eta W_1 \boldsymbol{x}^\mu P \left( \boldsymbol{y}^\mu - W_2 W_1 \boldsymbol{x}^\mu \right), \\
    \Delta P^\mu &= \eta P \boldsymbol{y}^\mu \boldsymbol{y}^{\mu \top} \left( I - P^\top P \right) .
\end{aligned}
\label{eq:DeltaQ_P_mu}
\end{equation}
Summing over all \( p \) training examples, we obtain the average updates:
\begin{equation}
\begin{aligned}
    \Delta Q &= \frac{\eta}{p} \sum_{\mu=1}^{p} \Delta Q^\mu = \eta W_1 \left( \frac{1}{p} \sum_{\mu=1}^{p} \boldsymbol{x}^\mu \left( \boldsymbol{y}^\mu - W_2 W_1 \boldsymbol{x}^\mu \right)^\top  P^\top  \right), \\
    \Delta P &= \frac{\eta}{p} \sum_{\mu=1}^{p} \Delta P^\mu = \eta P  \left( \frac{1}{p} \sum_{\mu=1}^{p} \boldsymbol{y}^\mu \boldsymbol{y}^{\mu \top} \right) \left( I - P^\top P \right).
\end{aligned}
\end{equation}
We simplify the update for \( Q \):
\begin{align}
    \Delta Q &= \eta W_1 \left(\left( \frac{1}{p} \sum_{\mu=1}^{p} \boldsymbol{x}^\mu \left( \boldsymbol{y}^\mu - W_2 W_1 \boldsymbol{x}^\mu \right)^\top \right) P^\top \right)\\
    &= \eta W_1 \left(\left( \Sigma_{io}^\top - W_1^\top W_2^\top \Sigma_{ii} \right) P^\top \right) \\
    &= \eta W_1 \left( \left( \Sigma_{io} - W_2 W_1 \right)^\top P^\top \right)
\end{align}
Similarly, the update for \( P \) simplifies to:
\begin{align}
    \Delta P &= \eta P \Sigma_{oo} \left( I - P^\top P \right).
\end{align}
Thus, the updates for \( Q \) and \( P \) become:
\begin{equation}
\begin{aligned}
    \Delta Q &= \eta W_1 \left( \Sigma_{io} - W_2 W_1 \right)^\top P^\top, \\
    \Delta P &= \eta P \Sigma_{oo} \left( I - P^\top P \right).
\end{aligned}
\end{equation}

Under the continuous-time assumption, where \( \eta \rightarrow 0 \) with \( \eta = \frac{dt}{\tau} \), the updates for \( Q \) and \( P \) become differential equations:
\begin{equation}
    \tau \frac{dQ}{dt} = W_1 \left( \Sigma_{io} - W_2 W_1 \right)^\top P^\top,
    \label{eq:Q_dynamics}
\end{equation}
and
\begin{equation}
    \tau \frac{dP}{dt} =  P \Sigma_{oo} \left( I - P^\top P \right).
    \label{eq:P_dynamics}
\end{equation}

Finally, substituting \( B = Q P \) into the update for \( W_1 \) from eq. \ref{eq:W1_dynamics}, we find that the dynamics for \( W_1 \) become:
\begin{equation}
    \tau \frac{d W_1}{dt} = Q P \left( \Sigma_{io} - W_2 W_1 \right)
    \label{eq:W1_dynamics_QP}
\end{equation}

We perform rotations similar to before:
\begin{equation}
\begin{aligned}
    W_1 &= \bar{W}_1 V^\top, \\
    W_2 &= U \bar{W}_2, \\
    P &= \bar{P} U^\top.
\end{aligned}
\end{equation}

Substituting for \( Q \) we get:
\begin{align}
    &\tau \frac{d Q}{dt} = \bar{W}_1 V_1^\top \left( USV^\top - U\bar{W}_2 \bar{W}_1 V^\top \right)^\top (PU^\top)^\top\\
    &= \bar{W}_1 V_1^\top V \left( S - \bar{W}_2 \bar{W}_1 \right)^\top U^\top U \bar{P}^\top \\
    &= \bar{W}_1  \left( S - \bar{W}_2 \bar{W}_1 \right)^\top \bar{P}^\top
\end{align}

Substituting these rotations into the previous derivations, we update the dynamics.

For \( W_1 \), the update equation is:
\begin{equation}
    \tau \frac{d W_1}{dt} = B \left( \Sigma_{io} - W_2 W_1 \right).
\end{equation}

Since \( B = Q P = Q \bar{P} U^\top \), \( W_1 = \bar{W}_1 V^\top \), \( W_2 = U \bar{W}_2 \), and \( \Sigma_{io} = U S V^\top \), we have:
\begin{align}
    \tau \frac{d (\bar{W}_1 V^\top)}{dt} &= Q \bar{P} U^\top \left( U S V^\top - U \bar{W}_2 \bar{W}_1 V^\top \right) \\
    &= Q \bar{P} \left( S V^\top - \bar{W}_2 \bar{W}_1 V^\top \right).
\end{align}

Since \( V^\top \) is constant, we can write:
\begin{equation}
    \tau \frac{d \bar{W}_1}{dt} V^\top = Q \bar{P} \left( S V^\top - \bar{W}_2 \bar{W}_1 V^\top \right).
\end{equation}

Multiplying both sides on the right by \( V \) (using \( V^\top V = I \)):
\begin{equation}
    \tau \frac{d \bar{W}_1}{dt} = Q \bar{P} \left( S - \bar{W}_2 \bar{W}_1 \right).
    \label{eq:W1_rotated_dynamics_QP}
\end{equation}

Substituting \( P = \bar{P} U^\top \) and \( \Sigma_{oo} = U S^2 U^\top \) in \eqref{eq:P_dynamics}, we get:
\begin{align}
    \tau \frac{d (\bar{P} U^\top)}{dt} &=  \bar{P} U^\top \left( U S^2 U^\top \right) \left( I - U \bar{P}^\top \bar{P} U^\top \right) \\
    &=  \bar{P} S^2 \left( I - \bar{P}^\top \bar{P} \right) U^\top.
\end{align}

Multiplying both sides on the right by \( U \) (since \( U^\top U = I \)):
\begin{equation}
    \tau \frac{d \bar{P}}{dt} = \bar{P} S^2 \left( I - \bar{P}^\top \bar{P} \right).
    \label{eq:P_rotated_dynamics}
\end{equation}

In summary, under the rotations, the updated dynamics are:
\begin{equation}
\begin{aligned}
    \tau \frac{d \bar{W}_1}{dt} &= Q \bar{P} \left( S - \bar{W}_2 \bar{W}_1 \right), \\
    \tau \frac{d \bar{W}_2}{dt} &= \left( S - \bar{W}_2 \bar{W}_1 \right) \bar{W}_1^\top, \\
    \tau \frac{d \bar{P}}{dt} &= \bar{P} S^2 \left( I - \bar{P}^\top \bar{P} \right). \\
    \tau \frac{d Q}{dt} &= \bar{W}_1  \left( S - \bar{W}_2 \bar{W}_1 \right)^\top \bar{P}^\top \\
\end{aligned}
\end{equation}

\section{Learning rule implementation}
Our implementation and optimization were conducted entirely using PyTorch. For the LDFA layers, we initialized all weights using Kaiming uniform initialization. We modified the backward pass by adjusting the gradients with respect to the input, ensuring they align with our proposed update rule as well as calculating the gradients of $P$ and $Q$.

We implemented two variants of the algorithm. In the **Normative** version (Algorithm 1), $P$ and $Q$ are updated to explicitly minimize the reconstruction error $\|QP - W^\top\|$, ensuring the feedback pathway approximates the transpose of the forward weights. In the **Local** version (Algorithm 2), $P$ and $Q$ are updated using biologically plausible, local plasticity rules (Oja's rule) that align the feedback subspace with the principal components of the error signal.

\begin{algorithm}[H]
\caption{Backward Pass for Normative LDFA (Approximating $W^\top$)}
\label{algorithm:Normative_LDFA}
\KwIn{Error signal $\boldsymbol{\delta}_{l+1} \in \mathbb{R}^{b \times m}$, activations $\boldsymbol{h}_l \in \mathbb{R}^{b \times n}$, matrices $Q_l \in \mathbb{R}^{n \times r}$,\\
\hspace{2.8em} $P_l \in \mathbb{R}^{r \times m}$, weight $W_l \in \mathbb{R}^{m \times n}$}
\KwOut{Gradient w.r.t.\ input $\text{grad\_input}$, updates $\Delta Q_l$, $\Delta P_l$, $\Delta W_l$}

\BlankLine
\textbf{Compute 'gradient' w.r.t.\ input:}\\
$\text{grad\_input} \leftarrow \left( Q_l P_l \boldsymbol{\delta}_{l+1}^\top \right)^\top \in \mathbb{R}^{b \times n}$\;

\textbf{Compute update for $W_l$:}\\
$\Delta W_l \leftarrow \boldsymbol{\delta}_{l+1}^\top \boldsymbol{h}_l$\;

\textbf{Compute updates for $Q_l$ and $P_l$ (Reconstruction):}\\
$E \leftarrow W_l^\top - Q_l P_l$\;
$\Delta Q_l \leftarrow E P_l^\top$\;
$\Delta P_l \leftarrow Q_l^\top E$\;

\end{algorithm}

\vspace{1em}
In the local version (Algorithm 2), we learn the projection matrix $\boldsymbol{P}$ to capture the principal directions of the target labels $\boldsymbol{y}$ (in the output layer) or the error signal $\boldsymbol{\delta}_{l+1}$ (in hidden layers). Specifically, for a layer $l$ with input dimension $n$, output dimension $m$, and rank constraint $r$, we proceed as follows:

\begin{algorithm}[H]
\caption{Backward Pass for Local LDFA (Oja's Rule)}
\label{algoirthm:Local_LDFA}
\KwIn{Error signal $\boldsymbol{\delta}_{l+1} \in \mathbb{R}^{b \times m}$, activations $\boldsymbol{h}_l \in \mathbb{R}^{b \times n}$, matrices $Q_l \in \mathbb{R}^{n \times r}$,\\
\hspace{2.8em} $P_l \in \mathbb{R}^{r \times m}$, weight $W_l \in \mathbb{R}^{m \times n}$}
\KwOut{Gradient w.r.t.\ input $\text{grad\_input}$, updates $\Delta Q_l$, $\Delta P_l$, $\Delta W_l$}

\BlankLine
\textbf{Compute 'gradient' w.r.t.\ input:}\\
$\text{grad\_input} \leftarrow \left( Q_l P_l \boldsymbol{\delta}_{l+1}^\top \right)^\top \in \mathbb{R}^{b \times n}$\;

\textbf{Compute update for $W_l$:}\\
$\Delta W_l \leftarrow \boldsymbol{\delta}_{l+1}^\top \boldsymbol{h}_l$\;

\textbf{Compute covariance matrix for P update:}

\If{use\_targets}{
    $C \leftarrow (\boldsymbol{y} - \boldsymbol{\bar{y}})^\top (\boldsymbol{y} - \boldsymbol{\bar{y}})$\;
}
\Else{
    $C \leftarrow (\boldsymbol{\delta_{l+1}} - \boldsymbol{\bar{\delta_{l+1}}})^\top (\boldsymbol{\delta_{l+1}} - \boldsymbol{\bar{\delta_{l+1}}})$\;
}
$\gamma \leftarrow \displaystyle \max \left( \operatorname{diag} \left( C \right) \right)$\;

\textbf{Compute updates for $Q_l$ and $P_l$ (Local):}\\
$\Delta Q_l \leftarrow \boldsymbol{h}_l^\top \left( P_l \boldsymbol{\delta}_{l+1}^\top \right)^\top$\;

$\Delta P_l \leftarrow P_l \left( \dfrac{C}{\gamma} \right) \left( I - P_l^\top P_l \right)$\;

\end{algorithm}

The derivation of the dLDFA (directed restricted adaptive feedback) follows the same procedure as Algorithm \ref{algoirthm:Local_LDFA}, with the key difference being that the error signal $\boldsymbol{\delta}$ does not necessarily originate from the next layer; instead, it can come from any subsequent layer.

\section{Numerical experiments}

\subsection{Common Architectures and Training Details}
Across our experiments, we utilized three primary architectures: a fully connected Multi-Layer Perceptron (MLP), a VGG-style Convolutional Neural Network (CNN), and a Vision Transformer (ViT). Unless otherwise stated in the specific experiment subsections, the architectures follow the specifications below.

\subsubsection*{Multi-Layer Perceptron (MLP)}
For the dense network experiments on CIFAR-10 and CIFAR-100, we used a network with 4 hidden layers.
\begin{itemize}
    \item \textbf{Structure:} Input $\to$ [Linear(512) $\to$ ReLU] $\times$ 3 $\to$ Linear(Output).
    \item \textbf{Hidden Width:} 512 neurons (unless specified otherwise for scaling experiments).
    \item \textbf{Initialization:} Kaiming Uniform.
\end{itemize}

\subsubsection*{Convolutional Neural Network (VGG-Like)}
We trained a VGG-style network with four convolutional blocks and batch normalization. The detailed architecture is provided in Table \ref{tab:cnn_arch}.

\begin{table}[H]
\centering
\caption{VGG-style CNN Architecture used for CIFAR-10 experiments.}
\label{tab:cnn_arch}
\small
\begin{tabular}{l|l|l}
\toprule
\textbf{Block} & \textbf{Layer Type} & \textbf{Details} \\
\midrule
\multirow{3}{*}{Block 1} 
 & $2 \times$ (Conv2d + BN + ReLU) & $3\times3$ kernel, 64 filters, padding=1 \\
 & MaxPool2d & $2\times2$ kernel, stride=2 \\
 & Output Size & $16 \times 16 \times 64$ \\
\midrule
\multirow{3}{*}{Block 2} 
 & $2 \times$ (Conv2d + BN + ReLU) & $3\times3$ kernel, 128 filters, padding=1 \\
 & MaxPool2d & $2\times2$ kernel, stride=2 \\
 & Output Size & $8 \times 8 \times 128$ \\
\midrule
\multirow{3}{*}{Block 3} 
 & $2 \times$ (Conv2d + BN + ReLU) & $3\times3$ kernel, 256 filters, padding=1 \\
 & MaxPool2d & $2\times2$ kernel, stride=2 \\
 & Output Size & $4 \times 4 \times 256$ \\
\midrule
\multirow{3}{*}{Block 4} 
 & $2 \times$ (Conv2d + BN + ReLU) & $3\times3$ kernel, 512 filters, padding=1 \\
 & AdaptiveAvgPool2d & Output size $(1, 1)$ \\
 & Flatten & Vector size 512 \\
\midrule
\multirow{3}{*}{Classifier} 
 & Linear + ReLU & $512 \to 256$ \\
 & Dropout & $p=0.4$ \\
 & Linear & $256 \to \text{Num Classes}$ \\
\bottomrule
\end{tabular}
\end{table}

\subsubsection*{Vision Transformer (ViT)}
We utilized the Vision Transformer implementation from the \texttt{timm} library. The default configuration used across experiments (unless explicitly modified for scaling tests) is detailed in Table \ref{tab:vit_params}.

\begin{table}[H]
\centering
\caption{Vision Transformer (ViT) hyperparameters.}
\label{tab:vit_params}
\begin{tabular}{ll}
\toprule
\textbf{Parameter} & \textbf{Value} \\
\midrule
Patch Size & 4 \\
Input Channels & 3 \\
Embedding Dimension & 384 (variable in scaling experiments) \\
Depth & 8 layers \\
Number of Heads & 8 \\
MLP Ratio & 2.0 \\
QKV Bias & True \\
Drop Rate & 0.1 \\
Attention Drop Rate & 0.1 \\
Drop Path Rate & 0.1 \\
\bottomrule
\end{tabular}
\end{table}

\subsection{Experiments with Normative Feedback}
In this section, we detail the experiments corresponding to Section 4 of the main text, where the feedback matrices $B_l = Q_l P_l$ are trained to minimize the reconstruction error $\|Q_l P_l - W_l^\top\|_F^2$ (Algorithm \ref{algorithm:Normative_LDFA}).

\paragraph{Dual Optimization Strategy} In all normative experiments, we employed \textbf{two separate optimizers}: one for the forward weights ($W_l$) and one for the backward feedback factors ($Q_l, P_l$). Both used the AdamW algorithm but were configured with different learning rates and schedules to ensure the feedback pathway effectively tracked the evolving forward weights.

\subsubsection*{Convolutional Networks (Layer-wise Constraints)}
To evaluate the impact of error compression on deep convolutional networks (Fig. 3a), we trained the VGG-like architecture defined in Table \ref{tab:cnn_arch} on the CIFAR-10 dataset.
\begin{itemize}
    \item \textbf{Rank Constraints:} We replaced the standard backward pass in every convolutional layer with a low-rank feedback channel. The rank $r$ was set to a fixed fraction of the layer's channel width $n$ (i.e., $r \in \{n/2, n/4, n/8\}$).
    \item \textbf{Feedback Kernel:} For convolutional layers, the projection $P_l$ operates as a $1 \times 1$ convolution, reducing the error signal from $m$ channels to $r$ channels per spatial location.
\end{itemize}

\subsubsection*{Vision Transformers (ViT)}
We trained the Vision Transformer using the dual-optimizer setup described below. The training duration was set to \textbf{250 epochs for CIFAR-10} and \textbf{300 epochs for CIFAR-100}.
\begin{itemize}
    \item \textbf{Scope:} Low-rank feedback was applied to all learnable linear transformations, including Attention (Q, K, V) and MLP layers.
    \item \textbf{Forward Optimization:} The forward weights were optimized using AdamW with a base learning rate of $3 \times 10^{-4}$ and weight decay of $0.1$.
    \item \textbf{Backward Optimization:} The feedback factors were optimized using AdamW with a base learning rate of $0.015$ and \textbf{no weight decay}.
    \item \textbf{Scheduling:} Both optimizers utilized a warmup phase of 10 epochs with a start factor of 0.1 (via linear scheduling), followed by a cosine annealing schedule decaying to a minimum learning rate of $1 \times 10^{-9}$.
    \item \textbf{CIFAR-100 Adaptation:} For CIFAR-100 experiments, the backward optimizer's base learning rate was adjusted to $0.005$.
\end{itemize}

\subsection{Experiments with Local Feedback}
In this section, we verify the efficacy of the local learning rule (Algorithm~\ref{algoirthm:Local_LDFA}), where the feedback projection $\boldsymbol{P}_l$ is learned via an error-driven Oja's rule to capture the principal subspace of the teaching signal.

\subsubsection*{Multi-Layer Perceptron (MLP) Experiments}
\paragraph{Experimental Setup}
Unless otherwise noted, we used the standard MLP architecture described in the Common Architectures section (4 hidden layers, 512 neurons). All models were trained for 160 epochs using the Adam optimizer with AMSGrad, a batch size of 32, a learning rate of $6 \times 10^{-4}$, and a weight decay of $4 \times 10^{-4}$. The learning rate followed an exponential decay with a factor of 0.975.

\paragraph{Layer-Wise Constraints}
To analyze the sensitivity of different depths to error compression, we applied the LDFA rank constraint to only one layer at a time, leaving all other layers unconstrained. We compared these distinct training runs against a baseline trained with standard backpropagation. Each experiment was repeated 10 times.

\paragraph{Constraining All Layers}
We evaluated the performance when the entire error pathway is compressed by applying rank constraints simultaneously to all layers with ranks $r \in \{64, 32, 16, 10\}$.
To verify that the network relies on high-dimensional feedforward representations even when feedback is low-rank, we trained a control model where the feedforward width was physically reduced to 64 neurons (matching the feedback bottleneck) without rank constraints. These experiments were repeated 10 times.

\paragraph{Task Dimensionality (CIFAR-100)}
To test the hypothesis that feedback rank scales with task dimensionality $d$, we trained the model on CIFAR-100 subsets with $d \in \{50, 75, 100\}$ classes. For each subset, we applied rank constraints to all layers. Each run was repeated 5 times.

\paragraph{Direct LDFA (dLDFA)}
We evaluated the local rule's ability to support broadcast error signals in two configurations:
\begin{itemize}
    \item \textbf{Output Broadcast:} The error signal $\boldsymbol{\delta}_L$ was propagated directly from the output layer to all preceding hidden layers.
    \item \textbf{Penultimate Broadcast:} The error signal was propagated from the penultimate layer to all earlier layers (applying rank constraints), while the final layer was updated normally.
\end{itemize}
These runs were repeated 5 times.

\subsubsection*{Convolutional Neural Networks (CNN)}
\paragraph{Methodology}
To extend LDFA to convolutional layers, we treat the rank constraint as a bottleneck on the number of error channels. The projection matrix $\boldsymbol{P}$ functions as a $1 \times 1$ convolutional filter, projecting the error signal at each spatial location from $m$ channels down to $r$ channels.
Crucially, the covariance matrix $C$ for Oja's rule (Algorithm~\ref{algoirthm:Local_LDFA}) is computed by aggregating over both the batch and spatial dimensions (i.e., across all pixels in all images). This ensures $\boldsymbol{P}$ captures the covariance structure of the pixel-wise error representations.

\paragraph{VGG-like Experiments}
We trained the VGG-style architecture defined in Table~\ref{tab:cnn_arch}. Training was conducted for 250 epochs using Adam with a learning rate of $5 \times 10^{-4}$, weight decay of $5 \times 10^{-5}$, and an exponential decay factor of 0.98. All experiments were repeated 5 times.

\subsubsection*{Neural Receptive Fields}
To analyze neural receptive fields, we trained the same convolutional network architecture as used in Lindsey et. al. \cite{lindsey2019unified}. The model consists of two convolutional layers with ReLU activations to simulate retinal processing, followed by three additional convolutional layers with ReLU activations to model the ventral visual stream (VVS). Fully connected layers were used for classification, with the complete architecture detailed in Table~\ref{table:RetinaModel}. A bottleneck was introduced between the retinal and VVS layers to constrain the information flow.

Training was performed using the low-dimensional feedback alignment (LDFA) method, where rank constraints were applied to feedback sent to the retina with ranks $r = 2, 4, 32$. The network was trained with hyperparameters similar to those in Linsdey et. al.~\cite{lindsey2019unified}, using the RMSProp optimizer with a learning rate of $1 \times 10^{-4}$, a weight decay of $1 \times 10^{-5}$, and an exponential learning rate decay factor of 0.985. Each model was trained for 120 epochs, and all experiments were repeated five times to ensure robustness.

Additionally, we conducted experiments with dynamic Restricted Adaptive Feedback (dLDFA), where feedback to the retina originated from higher visual layers while feedforward connections within the retina were constrained to four channels.



\begin{table}[h]
\centering
\caption{Retina model Architecture}
\label{table:RetinaModel}
\begin{tabular}{lcl}
\toprule
\textbf{Layer(s)} & \textbf{Output Size} & \textbf{Details} \\
\midrule
Input & $32 \times 32 \times 1$ & Grayscale input \\
\midrule
\multicolumn{3}{c}{\textbf{Retina}} \\
\midrule
Conv2D + ReLU & $32 \times 32 \times 32$ & $9 \times 9$ conv, 32 filters, padding=4 \\
Conv2D + ReLU & $32 \times 32 \times 32$ & $9 \times 9$ conv, 32 filters, padding=4 \\
\midrule
\multicolumn{3}{c}{\textbf{VVS}} \\
\midrule
Conv2D + ReLU & $32 \times 32 \times 32$ & $9 \times 9$ conv, 32 filters, padding=4 \\
Conv2D + ReLU & $32 \times 32 \times 32$ & $9 \times 9$ conv, 32 filters, padding=4 \\
Conv2D + ReLU & $32 \times 32 \times 32$ & $9 \times 9$ conv, 32 filters, padding=4 \\
\midrule
Flatten & $32,768$ & Flatten to vector \\
\midrule
\multicolumn{3}{c}{\textbf{Classifier}} \\
\midrule
Fully Connected + ReLU & $1,024$ & Linear layer, $32,768 \rightarrow 1,024$ \\
Dropout & $1,024$ & Dropout probability $p = 0.5$ \\
Fully Connected & $C$ & Linear layer, $1,024 \rightarrow C$ \\
\bottomrule
\end{tabular}
\end{table}

\end{appendices}

\end{document}